\documentclass[sigplan,nonacm,natbib=false]{acmart}
\pdfoutput=1

\AtBeginDocument{%
  }

\setcopyright{acmlicensed}
\copyrightyear{2018}
\acmYear{2018}
\acmDOI{XXXXXXX.XXXXXXX}

\acmConference[Conference acronym 'XX]{Make sure to enter the correct
  conference title from your rights confirmation emai}{June 03--05,
  2018}{Woodstock, NY}
\acmISBN{978-1-4503-XXXX-X/18/06}

\usepackage[firstinits=false,
style=ACM-Reference-Format,
uniquename=false,
uniquelist=false,
hyperref=auto,
maxbibnames=99,
maxcitenames=2,
style=numeric,
citestyle=numeric,
backref=false,
natbib=false,
backend=bibtex]{biblatex}

\AtEveryBibitem{%
	\clearfield{location}%
	\clearfield{volume}%
	\clearfield{pages}%
	\clearfield{number}%
	\clearfield{comment}%
	\clearfield{note}%
	\clearfield{abstract}%
	\clearfield{file}%
	\clearfield{organization}%
}

\usepackage[binary-units=true]{siunitx}
\usepackage[acronym]{glossaries}
\usepackage{caption}
\usepackage{subcaption}
\usepackage{xspace}
\usepackage{multirow}
\usepackage{multicol}
\usepackage{xcolor} 

\usepackage[frozencache]{minted}

\definecolor{LightGray}{gray}{0.9}

\newcolumntype{L}[1]{>{\raggedright\arraybackslash}m{#1}}
\newcolumntype{C}[1]{>{\centering\arraybackslash}m{#1}}
\newcolumntype{R}[1]{>{\raggedleft\arraybackslash}m{#1}}

\makeatletter
\newcommand\notsotiny{\@setfontsize\notsotiny\@vipt\@viipt}
\makeatother

\usepackage{siunitx}

\DeclareSIUnit{\fps}{fps}
\DeclareSIUnit{\loc}{LoC}

\usepackage{scalerel}
\usepackage{tikz}
\usetikzlibrary{svg.path}

\makeatletter
\let\MYcaption\@makecaption
\makeatother
\usepackage[font=footnotesize]{subcaption}
\makeatletter
\let\@makecaption\MYcaption
\makeatother

\newcommand{\dirquote}[1]{``#1''}
\newcommand{\eigName}[1]{``\textit{#1}''}

\newcommand*\circled[1]{\tikz[baseline=(char.base)]{
		\node[shape=circle,draw,inner sep=0.05pt] (char) {#1};}}

\usepackage[%
textsize=footnotesize
]{todonotes}
\usepackage{hyperref}
\AtBeginDocument{
	\hypersetup{
		pdftitle = {The Anatomy of a Triton Attention Kernel},
		pdfauthor = {Burkhard Ringlein, Jan Van Lunteren, Tom Parnell, Radu Stoica},
	} 
}

\newacronym{vm}{VM}{virtual machine}
\newacronym{ml}{ML}{machine learning}
\newacronym{nn}{NN}{neural network}
\newacronym{llm}{LLM}{Large Language Models}
\newacronym{sla}{SLA}{service level agreement}
\newacronym{ig}{IG}{information gain}
\newacronym{bo}{BO}{Bayesian Optimization}
\newacronym{bbo}{BBO}{black box optimization}
\newacronym{gemm}{GEMM}{General Matrix Multiply}
\newacronym{ir}{IR}{Intermediate Representation}
\newacronym{loc}{LoC}{Lines of Code}
\newacronym{ai}{AI}{Artificial Intelligence}
\newacronym{dsl}{DSL}{Domain-specific Language}
\newacronym{cisc}{CISC}{Complex Instruction Set Computer}
\newacronym{risc}{RISC}{Reduced Insturction Set Computer}
\newacronym{jit}{JIT}{Just-in-Time}
\newacronym{sota}{SoTA}{state-of-the-art}
\newacronym{numa}{NUMA}{Non-Uniform Memory Access}
\newacronym{ssm}{SSM}{state-space model}
\newacronym{ci}{CI}{Continous Integration}

\renewcommand{\baselinestretch}{0.976}
\makeatletter
\let\ACM@origbaselinestretch\baselinestretch
\makeatother
\begin{document}

\title{The Anatomy of a Triton Attention Kernel}
\subtitle{How to achieve cross-platform state-of-the-art LLM attention using only Triton}



%
%
%

\author{
	\href{https://orcid.org/0000-0002-7222-9539}{Burkhard Ringlein}, 
	\href{https://orcid.org/0009-0007-4110-9239}{Jan van Lunteren}, 
	\href{https://orcid.org/0009-0005-8089-866X}{Radu Stoica}, 
	\href{https://orcid.org/0000-0002-1308-6590}{Thomas Parnell}    \\          
	IBM Research \\ 
	Zurich, Switzerland \\
	{ngl, jvl, rst, tpa}@zurich.ibm.com
}

\renewcommand{\shortauthors}{Ringlein, van Lunteren et al.}

\newcommand{\version}{versions/v0}
\begin{abstract}	
A long-standing goal in both industry and academia is to develop an LLM inference platform that is portable across hardware architectures, eliminates the need for low-level hand-tuning, and still delivers best-in-class efficiency.
In this work, we demonstrate that portable, efficient cross-platform LLM inference is indeed possible and share our experience.
We develop a state-of-the-art paged attention kernel, the core performance-critical component of many LLM deployments, that builds exclusively on the domain-specific just-in-time compiled language Triton to achieve state-of-the-art performance on both NVIDIA and AMD GPUs. 
We describe our high-level approach, the key algorithmic and system-level improvements, the parameter auto-tuning required to unlock efficiency, and the integrations into a popular inference server that are necessary to bring the performance of a generic Triton attention kernel from 19.7\% of the state-of-the-art to 105.9\%.
Our results highlight how open-source domain-specific languages can be leveraged to unlock model portability across different GPU vendors.

\end{abstract}


	


\maketitle


\section{Introduction}
\begin{figure}[t]
	\centering
	\includegraphics[width=0.8\linewidth]{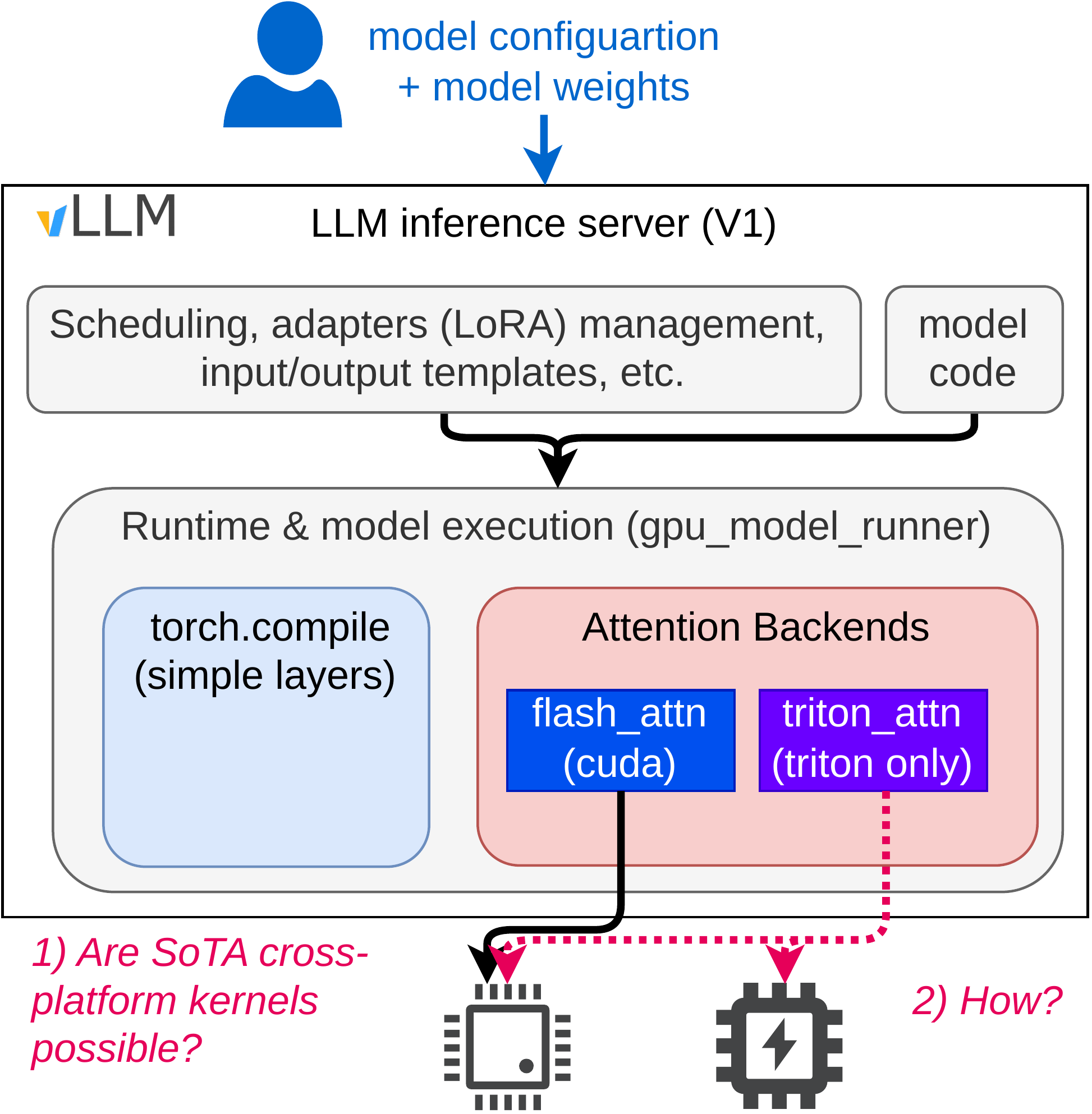}
	\vspace*{-1em}
	\caption{Architecture of vLLM v1 and the question of cross-vendor performance portability.}
	\label{fig:vllm-architecture}
	\vspace*{-1em}
\end{figure}

\glspl{llm} have evolved dramatically in the past years. Besides the improvement in model architectures and training procedures, there have been many innovations in optimizing LLM applications for modern hardware~\cite{ DaoFlashAttention2FasterAttention2023, ShahFlashAttention3FastAccurate2024, KwonEfficientMemoryManagement2023, YeFlashInferEfficientCustomizable2025, DaoFlashAttentionFastMemoryEfficient2022}.

However, the race in features and performance leads to a \dirquote{hardware lottery}~\cite{Hooker2021} for new \gls{ai} or \gls{ml} paradigms and to a gravity slope around the most dominant hardware platform. The tight interconnect between AI algorithms and AI hardware leads to limitations on the deployment and application scenario of \gls{ai}, since most features are only supported for a narrow set of hardware or input problem sizes~\cite{Hooker2021}. 
Consequently, the number and the size of libraries required to deploy \glspl{llm} with \gls{sota} performance have grown dramatically. 

While we welcome this Cambrian explosion of specialized libraries, this development adds an additional risk to democratizing \gls{ai}~\cite{Cohen2018}, since it creates a hard-to-navigate jungle of dependencies on frequently closed-source proprietary tools (e.g., \cite{YeFlashInferEfficientCustomizable2025}). 
The dependency on third-party libraries significantly hinders the adoption of new hardware and \gls{ai} applications. Additionally, writing tens of thousands of \gls{loc} to port a one-line kernel~\cite{VaswaniAttentionAllYou2017} slows down research and complicates deployment of new \gls{ml} methods unnecessarily.

A long-standing goal in both industry and academia is to develop a fully platform-independent software stack~\cite{Cohen2018}. At the inference server level, there is a tremendous momentum towards vLLM~\cite{KwonEfficientMemoryManagement2023, vllm}, an open-source framework that has the aspiration to be the de facto engine of \gls{llm} inference. 
vLLM abstracts away many of the complicated details of deploying modern LLMs, while being an active and vibrant open-source community. However, vLLM still depends on many proprietary or closed-source dependencies to compile and run LLM kernels, which are platform- and vendor-specific.
This dilemma is also depicted in \autoref{fig:vllm-architecture}. 
vLLM supports many platforms, however, for the performance-critical kernels,  it still depends on external libraries wrapped as \textit{backends} in vLLM. 
This is especially true for the \eigName{core} of many LLMs, the so-called (self-) attention layer \cite{VaswaniAttentionAllYou2017}, which is often the most performance-critical kernel of an LLM~\cite{ DaoFlashAttention2FasterAttention2023, ShahFlashAttention3FastAccurate2024, KwonEfficientMemoryManagement2023, YeFlashInferEfficientCustomizable2025, DaoFlashAttentionFastMemoryEfficient2022}. 
These platform-specific attention libraries have frequently tens of thousands of \gls{loc}~\cite{ringleinGPUPerformancePortability2025, ShahFlashAttention3FastAccurate2024}. 
%
%

This dilemma raises the important question, as also illustrated in the lower part of \autoref{fig:vllm-architecture}, if it is possible to design a fully platform-independent LLM implementation that can achieve state-of-the-art performance on multiple platforms? If so, what are the key steps required to develop a performance-critical, platform-independent kernel?
The aim of our research is to answer these questions.  

Recent efforts ~\cite{ringleinGPUPerformancePortability2025, dongFlexAttentionProgramming2024} indicate that it is possible to converge back to single-source open-source libraries, while combining all \gls{sota} LLM kernel innovations. 
In this work, we continue this line of research and describe and demonstrate the development of a production-ready, open-source, cross-platform, state-of-the-art LLM attention kernel. 
The approach we take is to build upon the OpenAI Triton domain-specific language (DSL) and show how the key steps required to make a cross-platform performance-critical kernel with SoTA performance.


We summarize our contributions as follows:
\begin{enumerate}
\vspace{-0.25em}
    \item We demonstrate a feature-complete cross-platform paged attention kernel with SoTA performance.
    \item We offer a comprehensive summary of the key lessons learned, including algorithmic optimizations, the programming and memory models adopted, and system-level trade-offs that are broadly applicable. 
	\item We open-source our kernels and micro-benchmark suite 
	(\href{https://ibm.biz/vllm-ibm-triton-lib}{ibm.biz/vllm-ibm-triton-lib}) and integrated the kernels into the one of the most widely used inference framework vLLM (\href{https://github.com/vllm-project/vllm}{vllm.ai}), where it has been adopted as the default attention kernel for AMD GPUs.
\end{enumerate}
 


\section{Background and Related Work}
\label{sec:related-work}

\subsection{vLLM: An Inference Server}
\label{subsec:vllm-background}

Typically, \glspl{llm} are deployed via serving or inference frameworks, such as vLLM~\cite{KwonEfficientMemoryManagement2023, vllm}, which abstract many details of the model deployment and request scheduling. 

\autoref{fig:vllm-architecture} shows an illustrative example of vLLM and its internal architecture, as of version 1 (\eigName{V1}). 
An inference server is necessary for reducing cost by allowing multiple users to use the same \gls{llm} in parallel and, therefore, are a critical part of the AI stack. Additionally, inference servers can decrease the latency and improve the throughput of the overall application, since the initial startup costs, such as loading model weights, are only incurred once. 
As shown at the top of \autoref{fig:vllm-architecture}, the user typically only provides the LLM to deploy, optional configuration regarding quantization, and the model weights, in case these are not publicly available in public repositories such as Hugging Face~\cite{HuggingFaceInc.HuggingFaceRepositories}. 
Today, vLLM is the de facto industry standard for LLM serving. vLLM is increasingly being adopted in production and can run on NVIDIA GPUs, AMD GPUs, as well as custom accelerators like AWS Inferentia~\cite{AWS2021}, Google's TPU~\cite{TPUArchitecture}, or IBM's Spyre~\cite{IntroducingIBMSpyre2021, VllmprojectVllmspyre2025}.


To allow such flexibility, vLLM has a complex internal structure to separate the functionality of scheduling, model pre- and post-processing, and runtime execution. 
To achieve state-of-the-art performance, the vLLM runtime execution is split into two major parts, as shown in the lower half of \autoref{fig:vllm-architecture}. The simpler layers of the deployed \gls{llm}, for example, normalization- or projection-layers, are written in platform-independent PyTorch functions and compiled and optimized automatically by using \texttt{torch.compile}~\cite{torchCompile2025}. However, the complex and most performance-critical attention layers are encapsulated in an abstraction called the attention \textit{backend}. In vLLM, there are multiple backends, usually wrappers around manually optimized libraries, such as \texttt{flash\_attn}~\cite{ShahFlashAttention3FastAccurate2024} or \texttt{flashinfer}~\cite{YeFlashInferEfficientCustomizable2025}, as shown in the middle of \autoref{fig:vllm-architecture}. Consequently, many backends depend on external libraries and vLLM cannot be used without them. 

\subsection{Triton: A tiling DSL}
\label{subsec:triton-background}

\begin{listing}[t]
	\begin{minted}
		[
		frame=lines,
		%framesep=2mm,
		%baselinestretch=1.2,
		%bgcolor=LightGray,
		fontsize=\scriptsize,
		linenos,
		]
		{python}
instance_id = tl.program_id(axis=0)
my_block_start = instance_id * BLOCK_SIZE
offsets = my_block_start + tl.arange(0, BLOCK_SIZE)
mem_mask = offsets < n_elements
x = tl.load(x_ptr + offsets, mask=mem_mask)
y = tl.load(y_ptr + offsets, mask=mem_mask)
result = x + y
tl.store(output_ptr + offsets, result, mask=mem_mask)
	\end{minted}
    \vspace*{-0.5em}
    \captionof{listing}{A simple vector add program in Triton, \texttt{BLOCK\_SIZE} is the global tiling size, determined manually or by autotuning.
    }
    \label{lst:triton-vector-add}
\end{listing}

The \gls{dsl} Triton~\cite{TilletTritonIntermediateLanguage2019, triton} has recently become popular as a promising open-source alternative to writing custom CUDA kernels. Triton (sometimes called \textit{OpenAI Triton}) enables writing and debugging kernels using high-level Python code, which can be compiled and executed on various GPU architectures. Triton kernels have been shown 
that they can be both highly performant and portable across different GPU platforms. For this reason, Triton is growing in popularity; it is used for many \gls{llm} stacks and is an integral part of \texttt{pytorch.compile}. 
 
Triton leverages a \gls{jit} compiler and builds on the idea of \textit{hierarchical tiles} to automate memory coalescing, shared memory allocation, and synchronization between threads~\cite{TilletTritonIntermediateLanguage2019}. Listing \ref{lst:triton-vector-add} shows a one-dimensional parallelized vector addition in Triton. Triton kernels can be fine-tuned for different workload sizes or target architectures using hyperparameters, also called \textit{kernel configurations}. For example, in Listing~\ref{lst:triton-vector-add}, \texttt{BLOCK\_SIZE} is a configuration parameter that influences the scheduling across the GPU cores.
However, in practice, Triton kernels also require hand optimizations for specific 
workloads and do not perform equally well across GPU platforms. 
To counter this, Triton kernels and their (compiler-) parameters can be autotuned.

\subsection{Portability and Kernel Parameter Autotuning}
\label{subsec:autotuning-background}

Autotuning is a complementary technique to compilation and can 
further increase performance-portability. It helps the compiler find an optimal, or nearly optimal, set of kernel configuration parameters through trial and error by leveraging microbenchmarks during or ahead of compilation. 
The main benefit of autotuning is that it avoids manually writing tens of thousands of highly optimized lines of code, as seen in the libraries in \autoref{subsec:attn-libs}. It is especially helpful when porting or deploying applications on new hardware. 
%
Autotuned kernels augment compilation with empirical performance tuning, generating and benchmarking a wide range of kernel variants to select the best-performing configuration for the target hardware and scenario. 
Autotuning reduces the parameter space a compiler needs to consider for a specific kernel compilation. 
This method can explore significantly more of the optimization space — often an order of magnitude more variants~\cite{Diamantopoulos2020b, Moreau2018, DaSilva2021} — leading to improved performance and better code specialization. 
Due to this, autotuning offers better portability to compile-time trade-offs compared to purely compiler-based approaches, which can lead to very long compilation times~\cite{Ringlein2022} or even be impossible, due to the complexity of the problem~\cite{ringleinGPUPerformancePortability2025}. 

In our previous work, we have shown that autotuning enables portability of Triton kernels for LLMs on GPUs~\cite{ringleinGPUPerformancePortability2025}. 
There, we compared a Triton implementation of Flash Attention v2 \cite{DaoFlashAttention2FasterAttention2023} with the \texttt{flash\_attn} library and the ROCM flash attention implementation and demonstrated that Triton can achieve comparable performance to the vendor-specific \gls{sota} libraries on both an NVIDIA A100 and an AMD MI250, using the same kernel code. 

Outside of academic literature, autotuning for \gls{llm} applications is leveraged, for example, by Pytorch Inductor. PyTorch Inductor is the tuning frontend for \texttt{torch.compile}, which is also a \gls{jit} compiler that can utilize Triton as its backend. Pytorch Inductor selects different algorithms by simply trying all sequentially, and recently added support for Triton kernels~\cite{pytorch-inductor-2025}.

However, autotuning Triton kernels to increase performance-portability still is barely or sub-optimally used in practice, primarily due to its large overhead. In \autoref{sec:autotuning}, we present our solution to overcome 
these limitations.


\subsection{Attention Kernels and Libraries}
\label{subsec:attn-libs}

Flash Attention (or \texttt{flash\_attn}) is a popular open-source library for the attention algorithm~\cite{DaoFlashAttention2FasterAttention2023, ShahFlashAttention3FastAccurate2024, DaoFlashAttentionFastMemoryEfficient2022, DaoAILabFlashattention2025}. It contains more than \SI{70000}{\loc}, mostly CUDA. FlashAttention is known for pioneering many algorithmic innovations, such as the so called \eigName{flash trick} to improve the memory locality of the attention algorithm~\cite{DaoFlashAttentionFastMemoryEfficient2022}. Flash Attention is mainly optimized to run on the latest NVIDIA GPUs. 
%
There exists also \texttt{RocmAttention}~\cite{ROCmFlashattention2025}, a fork of FlashAttention containing AMD-specific optimizations and cross-compiling the CUDA code using \texttt{hipify}. 

However, the first versions of flash attention led to high memory fragmentation as, for every request, sufficient memory must be reserved to store the maximum possible number of tokens that can be generated. 
Hence, \eigName{Paged Attention}~\cite{KwonEfficientMemoryManagement2023} developed a \textit{paged} version of flash attention, to leverage the concept of paging to reduce GPU memory consumption. The idea is only to reserve a small amount of memory, e.g., 16 tokens for new requests, in a data structure called a page. If the request generates more than 16 tokens, a new page is allocated. 
Paged attention improves the efficiency of inference servers and is a core feature of inference platforms like vLLM. Many other attention libraries like \texttt{flash\_attn} or \texttt{flashinfer} followed suit and added paged versions of their algorithms. 

Flashinfer is a partly open-source library containing \SI{51000}{} \gls{loc} and depends on many proprietary binary artifacts~\cite{YeFlashInferEfficientCustomizable2025}. It is written in CUDA and can only be deployed on recent NVIDIA hardware. 

Flex Attention~\cite{dongFlexAttentionProgramming2024} tries to reduce the number of attention libraries and to cover a broader range of attention varieties. It offers an optimized attention implementation with a customizable scoring method. That way, they can support different sliding windows, paged or not paged, changes to soft-capping, or different attention masks. 

Aiter~\cite{ROCmAiter2025} is a recent AMD-specific inference library with the goal of being a full wrapper for the different kernels optimized for the latest AMD GPUs. Within Aiter, kernels are written in various languages, including Triton, HIP, CK, and even assembly. There exists the option to use Aiter within vLLM for deployments on AMD GPUs. 

\section{Solution Overview}
\label{sec:overview}

\begin{figure}[t]
	\centering
	\includegraphics[width=0.85\linewidth]{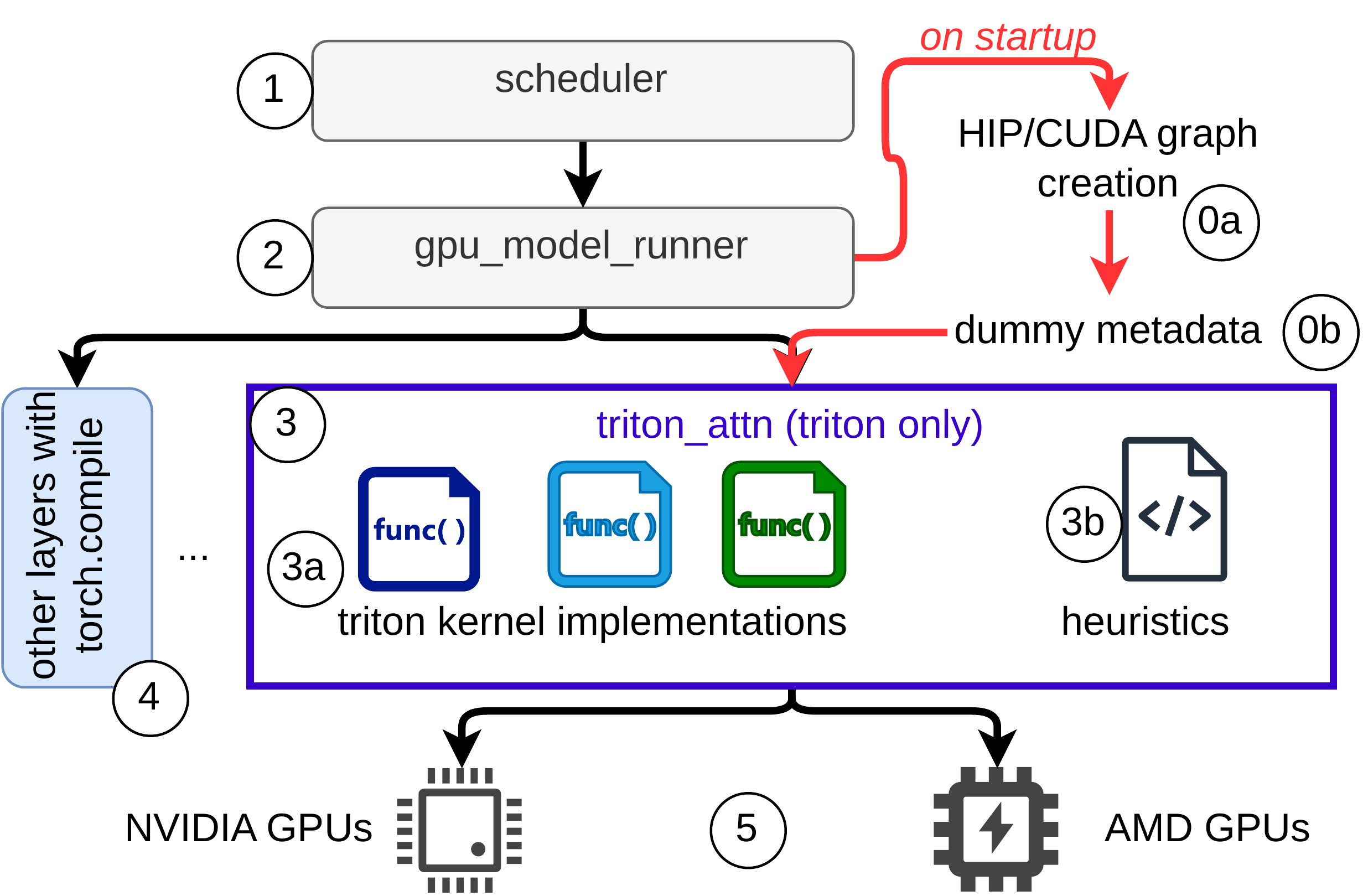}
	\vspace*{-1em}
	\caption{Overview of the \texttt{triton\_attn} backend in vLLM.}
	\label{fig:overview}
	\vspace*{-1em}
\end{figure}

An overview of our attention solution and its integration into vLLM is given in \autoref{fig:overview}. The figure shows all key steps and components that are relevant for achieving best-possible performance: The \texttt{scheduler} \circled{1} and \texttt{gpu\_model\_runner}  \circled{2} as vLLM core components, as well as our \texttt{triton\_attn} backend \circled{3} comprising three kernels \circled{3a}. Additionally, our backend includes configuration heuristics for these kernels \circled{3b} to improve performance portability.
The Triton Backend also takes into account on which GPU \circled{5} it is deployed when consulting these heuristics. 
In contrast to the manual optimized attention backends, most other layers in vLLM are written in native pytorch code and compiled to high-performance kernels using \texttt{torch.compile} \circled{4}.

A typical vLLM deployment records the CUDA-graphs or HIP-graphs at startup time \circled{0a}, before the inference server reports itself as ready. 
To do this, pseudo metadata are created \circled{0b} and feed to the model, including the attention backend. At inference time, these graphs are then \eigName{replayed} and the actual code is no longer executed. vLLM differentiates between two modes: \eigName{Partial CUDA-graphs} and \eigName{Full CUDA-graphs}. In the first mode, all the layers except attention and mamba layers are executed as CUDA graphs. In the latter mode, all layers are included in one CUDA graph recording and execution. 

In \autoref{sec:kernel-implementation}, we describe iteratively the development process of the \circled{3a} kernels. After, we describe how to tune the kernels and ensure best portability between vendors also using heuristics \circled{3b}, in \autoref{sec:autotuning}. Finally, we will describe the advantages and disadvantages of CUDA/HIP-graphs (\circled{0a} and \circled{0b}) and the implications of their usage in \autoref{sec:vllm-integration}.

\section{Design and Implementation of Attention} \label{sec:kernel-implementation}

\subsection{Core Concepts} \label{par:core_concepts}

Attention, as introduced in the seminal paper by Vaswani et
al.~\cite{VaswaniAttentionAllYou2017}, is computed according to:

\begin{equation}
	\text{Attention}(Q, K, V) = \text{softmax}\left(
	\frac{QK^\top}{\sqrt{d_k}} \right) V
	\label{eq:attention}
\end{equation}

with $Q, K, V \in \mathbb{R}^{n \times d_k}$  being the query, key, and value matrices,
respectively, derived through linear projections from the
input embeddings, with $n$ denoting the sequence length in tokens and
$d_k$ is the dimensionality of the key vectors. The intermediate matrix product $QK^\top \in \mathbb{R}^{n \times n}$
contains the attentions scores between all token pairs, while $\sqrt{d_k}$ is used as a
scaling factor for stabilizing the softmax operation.
A naive implementation of Equation~\ref{eq:attention} would incur a $\mathcal{O}(n^2)$ computation and memory complexity.
To accelerate the performance of attention computation on GPUs, various optimization techniques have been developed. These are briefly discussed below and
are also incorporated into our attention kernels, as discussed in this section.

\subsubsection*{Tiled Softmax}

The softmax function is calculated independently for each row in
$QK^\top$, where each row corresponds to one query token, to
produce a probability distribution by normalizing the attention
scores across all keys. It is computed in a numerically stable way
as follows:

\begin{equation}
	\text{softmax}(s_i) = \frac{e^{s_i - \max_j s_j}}{\sum_{k=1}^n
		e^{s_k - max_j s_j}}
\end{equation}

with $s_i$ denoting the $i$-th element of the score vector $s$
(i.e., a row in $QK^\top$), and $max_j s_j$ representing the
maximum value within that score vector. The latter is subtracted
from each element value to improve numerical stability by
preventing large positive values from causing overflow and large
negative values from prematurely underflowing to zero during
exponentiation.

The softmax can be computed efficiently using a tiled approach
(also denoted as online softmax) in which each row is partitioned
into smaller tiles that are processed incrementally,
rather than processing the entire row in a single pass. Tiled
softmax delays the division by the sum of exponentials to the end
of the computation. It maintains the maximum row value ($max_j
s_j$) and sum of exponentials ($\sum_{k=1}^m e^{s_k - max_j s_j}$)
and updates these after each tile is processed. This may involve a
rescaling of the intermediate results if the maximum changes.

The smaller tile sizes 
enables the kernel to use the fast shared memory and registers on the GPU most (or all) of the time, instead of falling back to the slower global memory. 
This results 
in substantial performance improvements. In practice, the tiled softmax
is fused with the tiled matrix multiplications in Equation~\ref{eq:attention}.
In the following, we will refer to this fused implementation as tiled softmax.
Tiled softmax is one of the key optimizations in
FlashAttention \cite{DaoFlashAttentionFastMemoryEfficient2022}.

\subsubsection*{KV Cache}

The computation of the attention layers of an \gls{llm} 
is accelerated by caching the $K$ and $V$
matrices for previously processed input tokens 
and reusing these cached matrices for subsequent
attention operations. The KV cache is initialized during prefill,
when $K$ and $V$ must be calculated for all tokens in the prompt.
During decode, however, K and V need only be calculated for each
newly generated token.

\subsubsection*{Grouped and Multi Query Attention}

State-of-the-art Transformer architectures are based on multi-head
attention, where multiple attention heads operate in parallel in
each layer, enabling the model to capture a broader range of
relationships. However, this comes at the cost of repeated
attention computation across heads, each having its own $Q$,
$K$ and $V$ projection matrices. Grouped Query Attention (GQA)
addresses this by reducing the number of key and value heads,
allowing multiple query heads to share the same $K$ and $V$
projections. This requires fewer K and V matrices to be computed
and results in a smaller KV cache. Multi Query Attention (MQA)
takes this to the extreme of using a single KV head for all
queries.

\subsubsection*{Batching}

Finally, and not limited to attention computation, batching multiple sequences
increases parallelism, which typically results in more
efficient utilization of GPU resources.

\subsection{Terminology}

We use the following terminology, which relates to a single sequence and is also used by vLLM: 

\begin{itemize}
	\item \textbf{Context Length:} The number of past tokens in the sequence whose $K$ and $V$ matrices are stored in the KV cache.
	\item \textbf{Query Length:} The number of new tokens that are currently being processed, for which attention is calculated against the cached context.
	\item \textbf{Sequence Length:} The total number of tokens in the sequence, defined as the sum of the context length and the query length.
\end{itemize}

For prefill attention, the context length equals zero, and the query length equals the prompt size. For decode attention, the query length equals one.

In addition, we use \textbf{Prefix Length} to denote the number of tokens preceding a given token in a sequence. This can include
both previously processed tokens and new tokens that appear before the current token (e.g., in a prompt).

\subsection{Baseline Triton Attention Kernel} \label{par:baseline}

\begin{figure}[t!]
  \centering
    \includegraphics[width=0.9\linewidth]{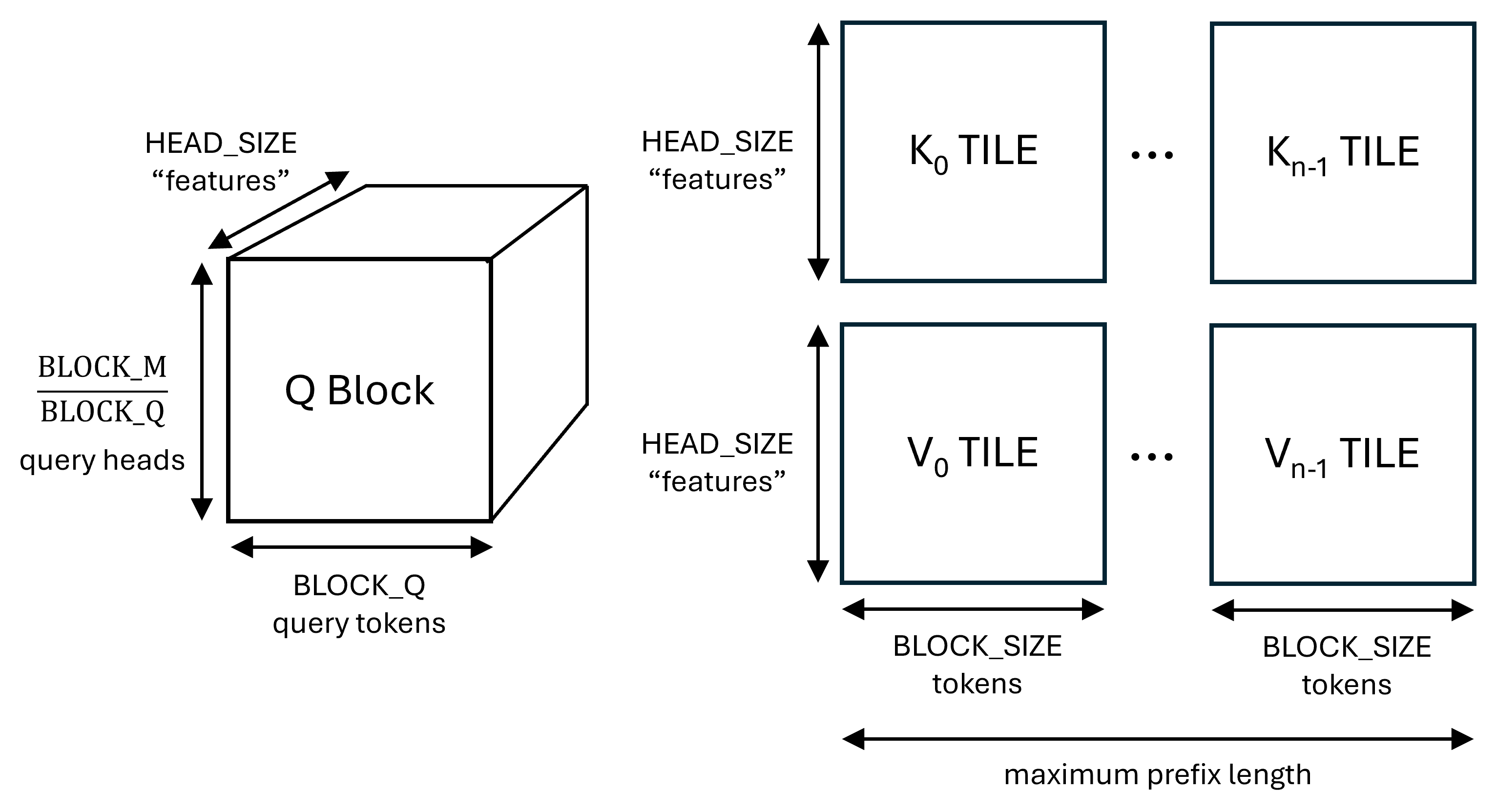}
    \vspace*{-1em}
    \caption{Q Block and KV Tiles.}
    \label{fig:q_block}
    \vspace*{-1em}
\end{figure}

For our first implementation of paged attention in Triton, we followed the original algorithm for paged attention~\cite{KwonEfficientMemoryManagement2023}.
Our implementation assumes that $Q$, $K$, and $V$ have already been computed before the kernel launch and stored in the KV cache in blocks, i.e. \eigName{paged memory}. The KV cache in vLLM is accessed through a block table (analogous to a page table), and the parameter \texttt{\small BLOCK\_SIZE} defines the maximum number of tokens stored in
a single KV cache block. 

Although we use the very same kernel for prefill and decode, we launch it with different launch grids for each phase. For the prefill phase, we launch \texttt{  tokens\_in\_batch} $\times$ \texttt{query\_heads} instances of the kernel, i.e., a two-dimensional launch grid. For the decode phase, we noticed that \texttt{  sequences\_ in\_batch} $\times$ \texttt{query\_heads} is a better launch grid.

We provide in \autoref{app:triton-baseline} a detailed description of the implementation, including Listing~\ref{lst:attention_baseline}.

\subsection{Prefill and GQA Optimization} \label{par:q_blocks}
The baseline attention kernel described in the previous section processes only a single pair of query token and query head per program instance.
We now introduce an optimization that increases the number of combinations handled within each program instance by including:

\begin{itemize}
	\item Multiple successive tokens from the same prompt, in case of prefill attention.
	\item Query heads that share the same KV head.
\end{itemize}

These combinations collectively form what we refer to as a Q Block, as illustrated in Figure~\ref{fig:q_block},
covering a total of \texttt{\small BLOCK\_M} combinations of \texttt{\small BLOCK\_Q} query tokens and $\frac{\texttt{BLOCK\_M}}{\texttt{BLOCK\_Q}}$ query heads.
By setting $\frac{\texttt{BLOCK\_M}}{\texttt{BLOCK\_Q}} = \frac{\texttt{num\_query\_heads}}{\texttt{num\_kv\_heads}}$, each Q Block cover all query heads that map to a single KV head.

Structurally, the Q Block is a three-dimensional block, with dimensions corresponding to
the number of query tokens, query heads, and the head size. However, for implementation efficiency, the Q Block is represented
as a two-dimensional tensor with a shape of $\texttt{\small BLOCK\_M}\ \times\ \texttt{\small HEAD\_SIZE}$.
This flattening simplifies memory access patterns and aligns better with Triton's programming model.

For a given sequence, a total of $\left\lceil \frac{\texttt{query\_length}}{\texttt{BLOCK\_Q}}\right\rceil$ Q Blocks are required to process all query tokens.
In the case of decode attention, where only a single query token is processed at a time (i.e., $\texttt{\small BLOCK\_Q} = 1$), this results in one Q Block per sequence.

The Q Block structure enables the kernel to process multiple attention computations in parallel, which improves efficiency.
For prefill attention, many query tokens attend to the same preceding tokens, allowing efficient reuse of the K and V matrices. Similarly, in Grouped Query Attention (GQA),
multiple query heads correspond to the same KV head, which only needs to be loaded once per Q Block. All of this reduces memory bandwidth and improves computational efficiency by increasing the arithmetic density.

The structure of $K$ and $V$ tiles processed for each Q Block using the tiled softmax approach remains unchanged from the baseline attention kernel
discussed in Section~\ref{par:baseline}, as also illustrated in Figure~\ref{fig:q_block}.
These tiles correspond to the KV head to which the query heads in the Q Block are mapped, and
span the tokens preceding those in the Q Block, up to the maximum prefix length of any token in the Q Block.
An example code for the Q-Block-based optimized attention kernel is described in \autoref{app:triton-qblock} and Listing~\ref{lst:attention_q_block}.

\subsection{Parallel Tiled Softmax} \label{par:par_tiled_softmax}
When using the kernels discussed in the previous sections for performing decode attention, the first dimension of the launch grid corresponds to the number of sequences in the batch.
As a result, small batch sizes lead to a limited number of program instances being launched, which can underutilize available GPU resources and result in degraded performance.
This limitation does not apply to prefill attention, because prompts typically contain many tokens, resulting in a sufficiently high number of Q Blocks to 
saturate 
the available compute resources on the GPU. 

\begin{figure}[t!]
  \centering
    \includegraphics[width=0.9\linewidth]{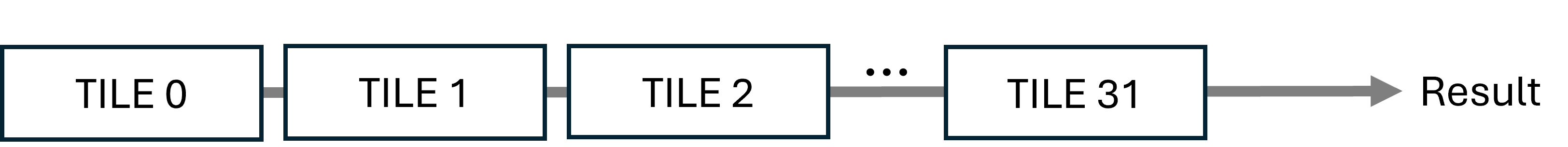}
    \vspace{2mm}
    (a)
    \vspace{2mm}
    
    \includegraphics[width=0.9\linewidth]{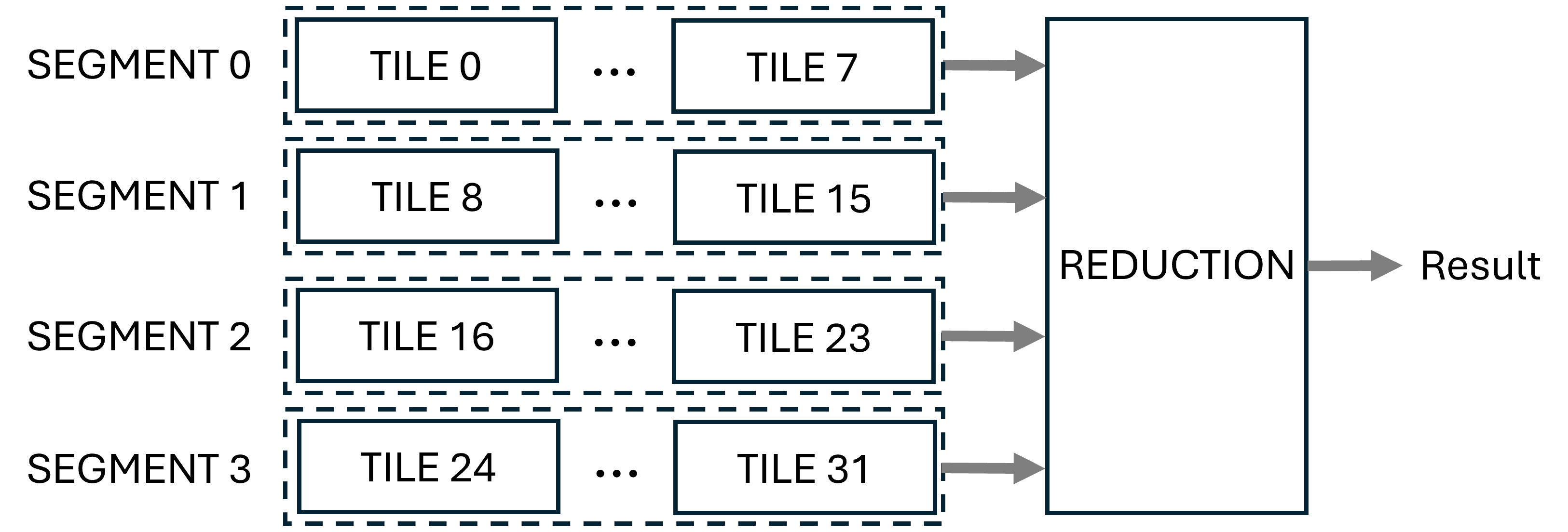}
    \vspace{2mm}
    (b)
    \vspace{-4mm}
    \caption{Parallel Tiled Softmax.}
    \label{fig:tiles_and_segments}
    \vspace*{-1em}
\end{figure}

To extract enough parallelism during decode attention, the iterative processing of tiled softmax can be parallelized across multiple program instances.
This approach is particularly advantageous for small batches of long sequences, where a large number of tiles must be processed and the GPU tends to be underutilized. By distributing the processing of these tiles across multiple program instances
executed in parallel, rather than executing them sequentially within one program instance, significant speedups can be achieved.
To describe our implementation of parallel tiled softmax, we introduce the following terminology. A \textit{tile} refers to the granularity at which the tiled softmax operations process $Q$ against $K$ and $V$ in an iterative fashion. A \textit{segment} comprises the tiles that are executed together within a single program
instance. This is illustrated by an example in Figure~\ref{fig:tiles_and_segments}. Figure~\ref{fig:tiles_and_segments}~(a) shows an attention computation partitioned into $32$ tiles, which are
processed iteratively while maintaining and updating intermediate results, maximum, and sum of exponentials, as discussed in Section~\ref{par:core_concepts}.
Figure~\ref{fig:tiles_and_segments}~(b) shows how these $32$ tiles are distributed over $4$ segments, each comprising $8$ tiles. The attention computation within the segments is performed
in the usual iterative manner, but in separate program instances. To produce the final result, a reduction step is required to merge the outputs from all segments. For this purpose,
the program instances executing the segments stores the intermediate results in memory. 
Finally, to obtain the final attention output, the intermediary results are retrieved, combined, and rescaled 
in a manner analogous to the iterative tiled softmax approach. 
An example implementation is presented in \autoref{app:triton-par-ts}.

The kernel implementing the parallel tiled softmax is only launched for decode attention on small batches involving longer sequences. This behavior is implemented by the heuristics in our Triton backend (\circled{3b} in \autoref{fig:overview}).
To determine whether to use this kernel or the original attention kernel, we apply a heuristic-based selection strategy.


\subsection{Adjustable Tile Sizes}
\label{par:adjustable-tiles}
\vspace*{-0.25em}

In the kernels discussed in the previous sections, the tile size used in the tiled softmax
was constrained to match the number of tokens contained in each KV cache block, defined by one parameter called \texttt{\small BLOCK\_SIZE}.
We have extended our kernel implementation to decouple the tile size from \texttt{\small BLOCK\_SIZE}, enabling it to be configured independently: smaller, equal, or larger. A key advantage is that this allows tuning the tile size independently of the block size and separately for prefill and decode for best performance. Another advantage is that it facilitates support for hybrid models that combine conventional Transformer attention layers with \gls{ssm} layers (e.g., Mamba~\cite{guMambaLinearTimeSequence2024}). In such architectures, attention layers often utilize large, non-power-of-two block sizes to achieve proper page alignment between the attention and \gls{ssm} layers.

\subsection{Static Launch Grid}
\label{par:static_launch_grid}

As the last optimization step, 
we modify the kernels 
to use a static launch grid, which consistently launches the same number of program instances. This was achieved by adapting the number of Q blocks processed by each instance. The primary benefit of this approach is improved compatibility with full CUDA or HIP graphs, as will be discussed in \autoref{subsec:cuda-graphs-vllm}.


\section{Autotuning and Portability}
\label{sec:autotuning}





\subsection{Disadvantages of Current Autotuning}

Recent works have shown that single-source Triton kernels can be competitive on multiple platforms, but may need autotuning (c.f. \autoref{subsec:autotuning-background}). However, in practice, autotuning of Triton kernels comes with a large overhead in tuning time~\cite{ringleinGPUPerformancePortability2025, RingleinRFCAutotunerDejavu2024, MaherCacheAutotuneTimings2025} which renders it unusable for many use cases, e.g., the use in vLLM (c.f. \cite{vllm_remove_tuning_KernelROCMUpstreamPrefix_2025}).
For example, tuning flash attention v2 extensively to achieve best-possible performance took nearly 24 hours for each GPU type~\cite{ringleinGPUPerformancePortability2025}.

This overhead can be reduced in some situations if the autotuning results are cached so they can be reused between deployments~\cite{ringleinGPUPerformancePortability2025, RingleinRFCAutotunerDejavu2024, MaherCacheAutotuneTimings2025, PyTorchCommunityPersistentCache2025, IBMTritondejavu2024}. These caches of the Triton autotuner contain the results of an autotuning run with a simple mapping of a scenario to the autotuning results in that scenario. A scenario is a specific combination of arguments passed to the Triton kernels, such as tensor pointers, their shapes, strides, and other scalar arguments. In vLLM attention backends, some of these kernel arguments change, e.g., with the sequence length. 
However, this caching of autotuning states helps only if the exact same scenario occurs again, in which case the autotuning phase is skipped and the cached result is used instead.
For example, after tuning for 32 tokens, a request might arrive again with 32 tokens; in this case, the autotuning of the attention kernel could be skipped. But if a request with 33 tokens arrives, then the kernel needs to be autotuned again.  

Besides the tuning overhead, autotuning in inference servers causes additional disadvantages: 
First, even if all possible scenarios are tuned ahead of time, the lookup of the optimal configuration to be used in a concrete scenario usually adds around 
tens of micro-seconds to the Triton kernel launch time, consuming the latency improvements of the tuned configuration for short workloads. 
Second, and more important, if using CUDA or HIP graphs, there is no way to look up the optimal configuration for each kernel launch, since HIP/CUDA graphs are recorded once and then just replayed without another consultation of the Triton autotuner (c.f. \autoref{sec:overview} and \autoref{subsec:cuda-graphs-vllm}).

\begin{figure}
	\centering
	\includegraphics[width=1.05\linewidth]{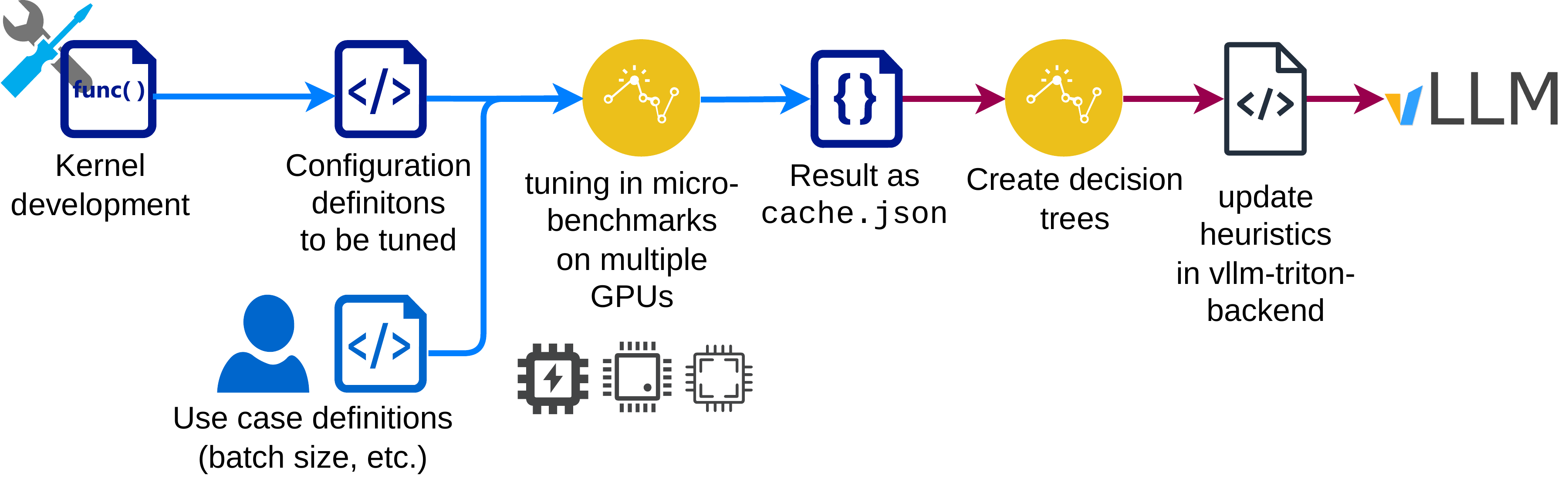}
	\vspace*{-1.5em}
	\caption{Workflow of tuning in vllm-triton-backend.}
	\label{fig:dejavu-flow}
	\vspace*{-1em}
\end{figure}

\subsection{Usage of autotuning in the vllm-triton-backend}


To circumvent the tuning time overhead and the inflexibility of the Triton autotuner, we decided to follow a two-step approach, as depicted in \autoref{fig:dejavu-flow}: 
First, we created a micro-benchmark framework to perform kernel tuning outside of the vLLM runtime components. This way, we avoid adding additional latency to the vLLM startup time or increasing the cost of \gls{ci} pipelines, where autotuning would be executed every time.
The micro-benchmarks are designed to call the same kernel code as the kernels in vLLM and simulate specific request patterns and \gls{llm} architectures. Additionally, the microbenchmarks help us better understand some performance artifacts of end-to-end results. 
For example, some kernels in the field are written for batches that always contain the same amount of tokens in every request, e.g., a batch of 8 prompts with 1000 tokens each. However, in reality, this is very unlikely for the most use cases of LLM inference serving. 
In our micro-benchmark framework, we can simulate varying context lengths, prompt lenghts, and batch sizes and adjust the kernels accordingly. 

After the kernel tuning using micro-benchmarks is completed, we analyze the results of the autotuning and export them as heuristics --- simple if-else \eigName{decision trees} in this case --- as a second step, as shown on the right-hand side of \autoref{fig:dejavu-flow}. 
One example of such a heuristic is given in Listing~\ref{lst:eval-tune-decision-tree}. 
In contrast to the static reuse of autotuning caches~\cite{ringleinGPUPerformancePortability2025, RingleinRFCAutotunerDejavu2024, MaherCacheAutotuneTimings2025, anyscaleHowIBMResearch2024,ringleinAchievingPlatformPortability2024}, decision trees have the advantage of 
providing optimized configurations 
for scenarios that are not in the autotuning cache, i.e., that were not part of the tuning.
Consequently, using simple decision trees has benefits at runtime and at tuning time: It also reduces the amount of tuning necessary, because tuning just for the average and corner use-cases will produce decision trees applicable to most scenarios. 




\section{Inference Server and System Integration}
\label{sec:vllm-integration}


Despite optimizing the kernels themselves to achieve \gls{sota} performance, our attention kernels need to be integrated into the vLLM. For this, we needed to adapt (1) the creation and computation of the attention metadata and (2) balance the trade-off of using HIP/CUDA graphs.

\subsection{Computation of Metadata}

In vLLM, after the scheduler decides which (partial) requests are included in the next batch to be processed, the data is copied to the GPU (if not already there), fitted into the paged data structures, and the corresponding metadata is computed. 
This attention metadata in vLLM contain, e.g., the tensors with the list of request lengths in the batch. For some backends, the batch is also sorted to start with decode or prefill requests. 

For the Triton backend, we needed to do the following adaptation: 
First, we count the number of decodes in the batch, so that we can decide whether the parallel tiled softmax version should be used or not.
Second, we construct a tensor that stores the accumulated number of Q Blocks for the sequences in the batch. In each launched program instance, this tensor is used to perform a binary search and determine the sequence index corresponding to the Q Block index for that instance, as described at the end of Section~\ref{par:q_blocks}. The total number of Q Blocks required for processing the batch is also derived from this tensor.

\subsection{Triton Launch Overhead and CUDA/HIP Graphs}
\label{subsec:cuda-graphs-vllm}

One major issue that must to be addressed for the integration of the Triton Backend is the use of CUDA~\cite{CUDAGraphManagement} or HIP~\cite{HIPGraphRuntimeAPI} graphs.
CUDA/HIP graphs are helpful to remove the software overhead of a forward pass through the model, if each forward pass invokes exactly the same computations. 
%
Triton kernels could benefit from individual configurations per batch shape, i.e., the use of different Triton configurations depending on the number of requests in the batch, their duration, etc. (c.f. \autoref{sec:autotuning}). 
However, this means that the Triton Attention Backend changes its behavior in every forward pass and cannot be executed with HIP/CUDA graphs.
In this approach, without CUDA/HIP graphs, the software overhead of Triton kernel launches is in the order of \SIrange{100}{300}{\micro\second}, according to our profiling. This launch overhead dominates the kernel execution time for sequences below roughly 1000 tokens (c.f. \autoref{subsec:microbench-eval}). 
This time excludes the \gls{jit} compilation time that occurs during the first kernel execution. Instead, it refers to the software overhead created at every kernel launch by Triton, e.g., by checking if another invocation of the JIT is required. 
Even if we circumvent some of these checks by caching them~\cite{JitCacheVllm2025}, we still have a launch overhead of around \SI{80}{\micro\second}.
%
To avoid this, we had to use CUDA/HIP graphs, but this creates a different set of trade-offs: For every CUDA/HIP graph, the arguments to all kernels in this graph are frozen~\cite{teamVLLMV1Major2025, CUDAGraphManagement, HIPGraphRuntimeAPI}.
This includes the pointers of tensors. Consequently, if the numbers of recorded CUDA/HIP graphs rise, the GPU memory gets filled by reserved memory for the CUDA/HIP graphs. Additionally, the CUDA/HIP graphs require some memory themselves. 
Therefore, vLLM decided to limit the number of graphs to one per batch size, and even only power-of-two batch sizes are considered~\cite{teamVLLMV1Major2025}. 

Hence, at vLLM startup time, all power-of-two batch sizes up to 128, usually, are recorded with one dummy request each. Since the required GPU memory for the kernels is allocated during this recording run, the recording run needs to occur with the maximum possible model sequence lengths. In other words, if using CUDA/HIP graphs, all kernels are always invoked as if the batch contains only requests of the maximum model length (or, depending on the implementation, maximum batched tokens). 
For our Triton kernels, this created another penalty, as we determine the number of kernel instances to launch based on the batch metadata, which is a common practice for Triton kernels. 
However, when using CUDA/HIP graphs, the launch grid is also fixed after the initial graph was recorded. 
Therefore, if we use CUDA/HIP graphs for our backend, we always launch as many Triton kernels as we would need for the longest request possible. If shorter requests are part of the batch, which is usually the case, the excess kernel instances will exit immediately; therefore, the computation of the kernel is always correct. But the launch of the excess instances still causes the GPU scheduler to schedule too many GPU \eigName{waves}, i.e., rounds of kernel executions for all GPU cores (i.e., Streaming Multiprocessors). 

Our evaluation revealed that the resulting additional runtime latency of the excess instances outweighs the saving of the launch overhead in nearly all cases. Therefore, we changed our kernel implementations to always have a static launch grid, close but smaller than the number of available GPU cores. 

\begin{listing}[t]
	\captionof{listing}{Decision tree as heuristic for \texttt{Triton (tuned)}.}
	\label{lst:eval-tune-decision-tree}
	\begin{minted}
		[
		frame=lines,
		%framesep=2mm,
		%baselinestretch=1.2,
		%bgcolor=LightGray,
		fontsize=\scriptsize,
		linenos,
		]
		{python}
BLOCK_M = 64 if max_seqlen_q > 1 and avg_seqlen_q >= 4096 \
                          and is_nvidia_gpu() \
                      else 16
BLOCK_N = 32 if max_seqlen_k <= 64 or avg_seqlen_q <= 4096 \
                         or is_amd_gpu() \ 
                     else 64
	\end{minted}
    \vspace*{-1em}
\end{listing}

%
%
%
%


\begin{figure*}[t]
	\vspace*{-1em}
	\centering
	\captionsetup[subfigure]{justification=centering}
	\begin{subfigure}[t]{.5\textwidth}
		\includegraphics[width=\linewidth]{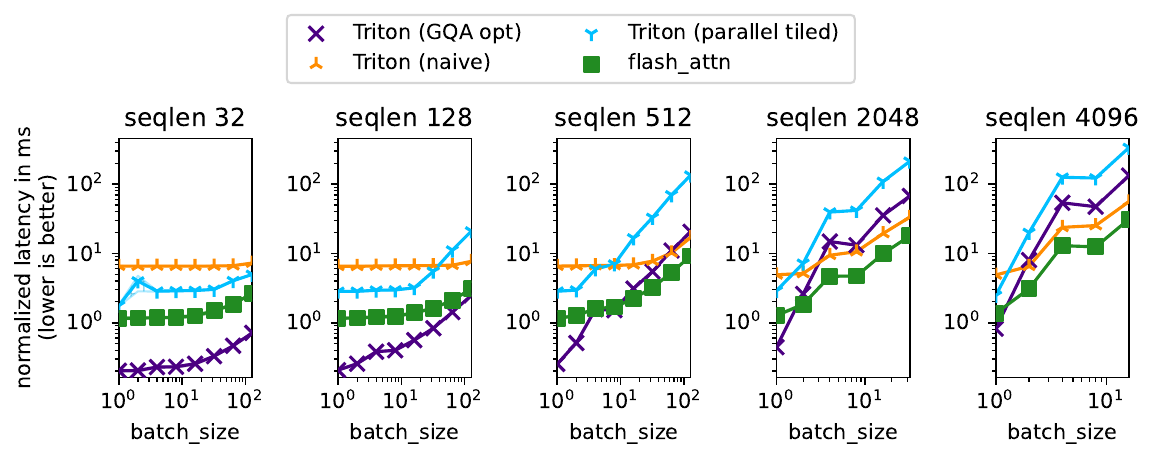}
		\vspace*{-2em}
		\caption{H100, sorted by max. sequence length}
		\label{subfig:eval-step1-h100}
	\end{subfigure}%
	\begin{subfigure}[t]{.5\textwidth}
		\includegraphics[width=\linewidth]{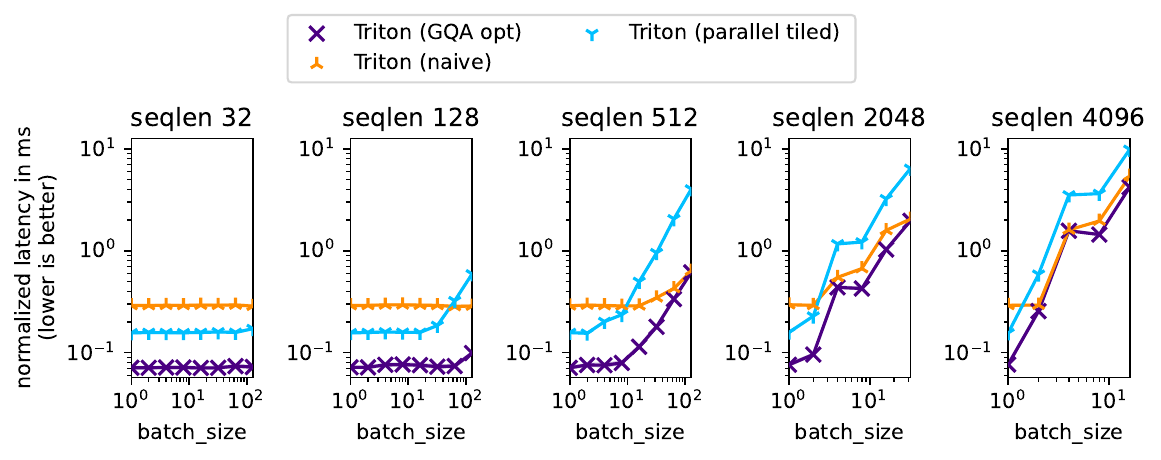}
		\vspace*{-2em}	
		\caption{MI300, sorted by max. sequence length}
		\label{subfig:eval-step1-mi300}
	\end{subfigure}%
	
		\begin{subfigure}[t]{.5\textwidth}
		\includegraphics[width=\linewidth]{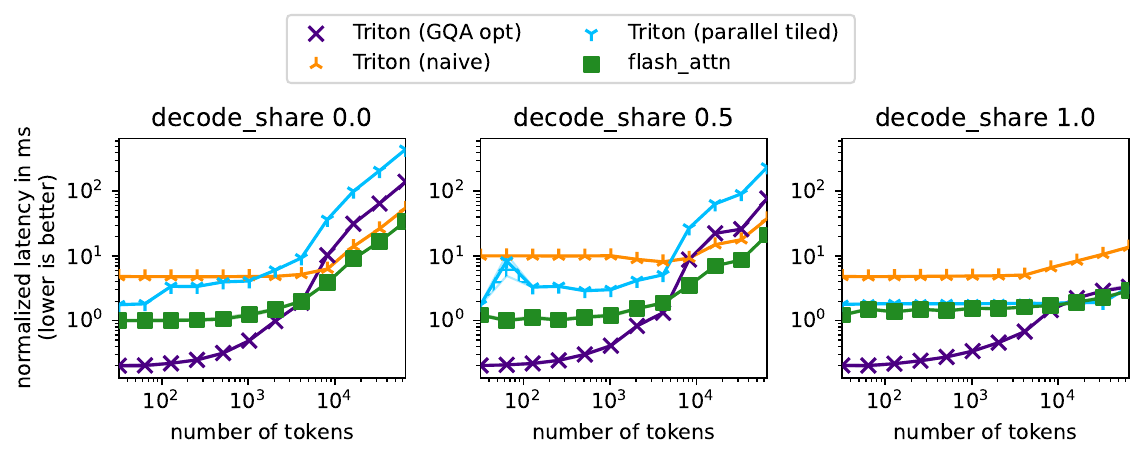}
		\vspace*{-2em}
		\caption{H100, sorted by share of decode requests}
		\label{subfig:eval-step1-h100-typewise}
	\end{subfigure}%
	\begin{subfigure}[t]{.5\textwidth}
		\includegraphics[width=\linewidth]{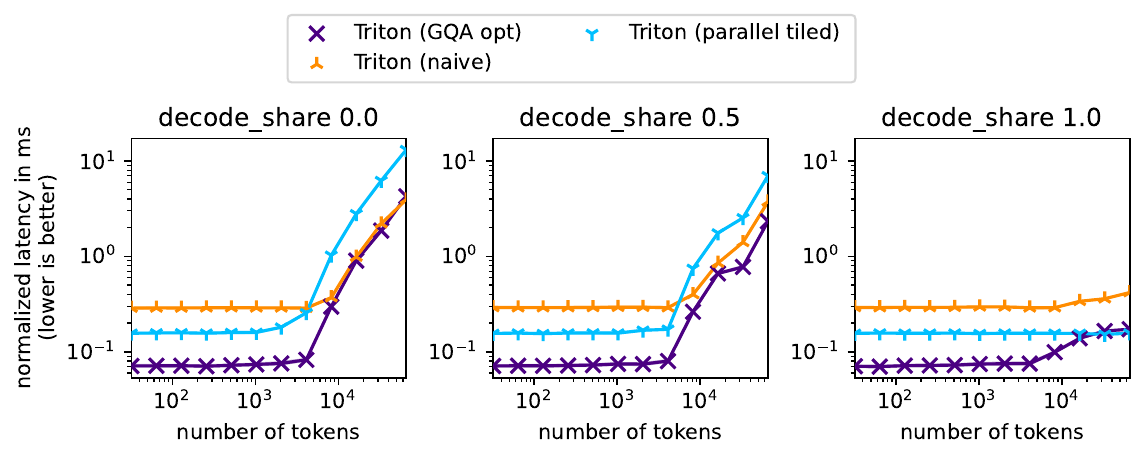}
		\vspace*{-2em}	
		\caption{MI300, sorted by share of decode requests}
		\label{subfig:eval-step1-mi300-typewise}
	\end{subfigure}%
	\vspace*{-0.5em}
	\caption{Comparing performance of different kernel optimizations with the baselines.
	}
	\label{fig:eval-step1}
	\vspace*{-1em}
\end{figure*}
\begin{figure*}[t]
	\centering
	\captionsetup[subfigure]{justification=centering}
	\begin{subfigure}[t]{.5\textwidth}
		\includegraphics[width=\linewidth]{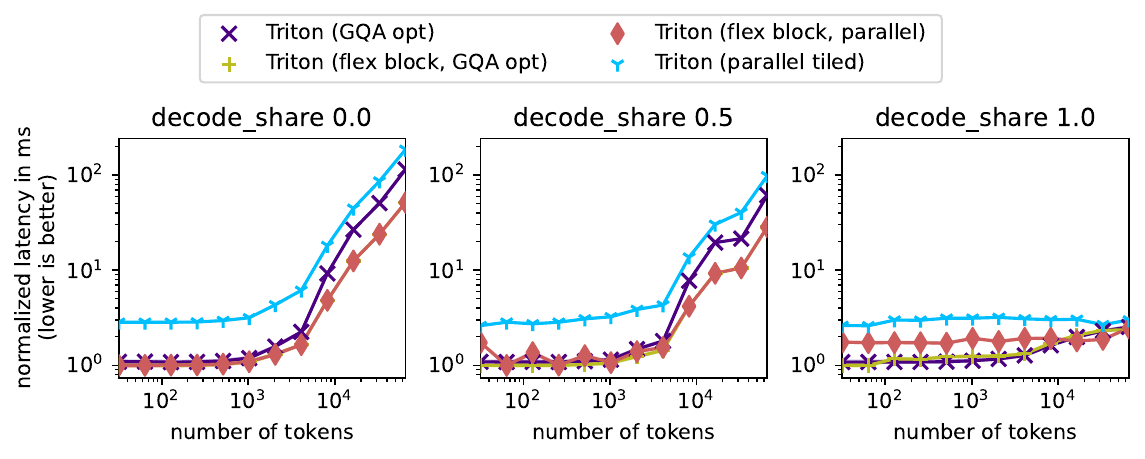}
		\vspace*{-2em}
		\caption{H100, sorted by share of decode requests}
		\label{subfig:eval-step2-h100}
	\end{subfigure}%
	\begin{subfigure}[t]{.5\textwidth}
		\includegraphics[width=\linewidth]{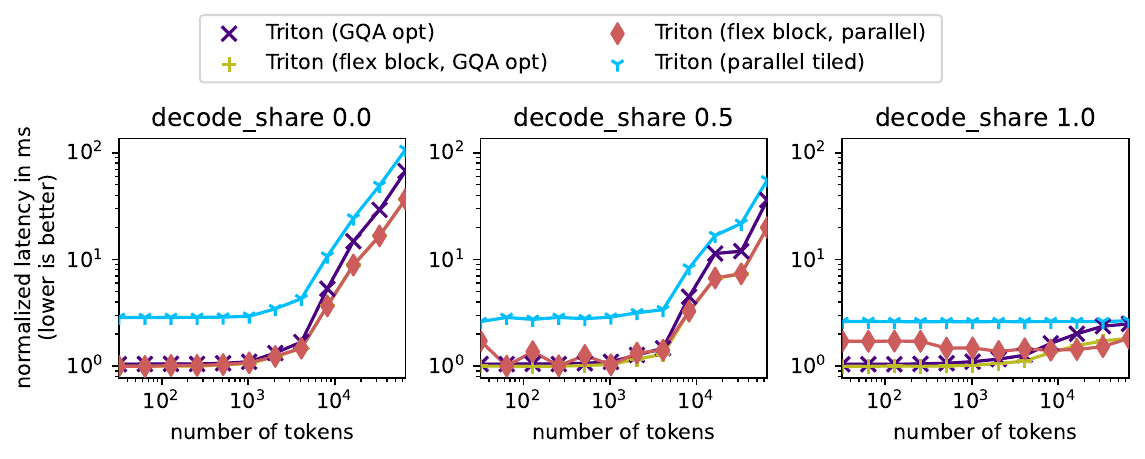}
		\vspace*{-2em}	
		\caption{MI300, sorted by share of decode requests}
		\label{subfig:eval-step2-mi300}
	\end{subfigure}%
	\vspace*{-0.5em}
	\caption{Comparing performance of the adjustable tile size optimization (c.f. \autoref{par:adjustable-tiles})
	}
	\label{fig:eval-step2}
	\vspace*{-1em}
\end{figure*}
\begin{figure*}[t]
	\centering
	\captionsetup[subfigure]{justification=centering}
	\begin{subfigure}[t]{.5\textwidth}
		\includegraphics[width=\linewidth]{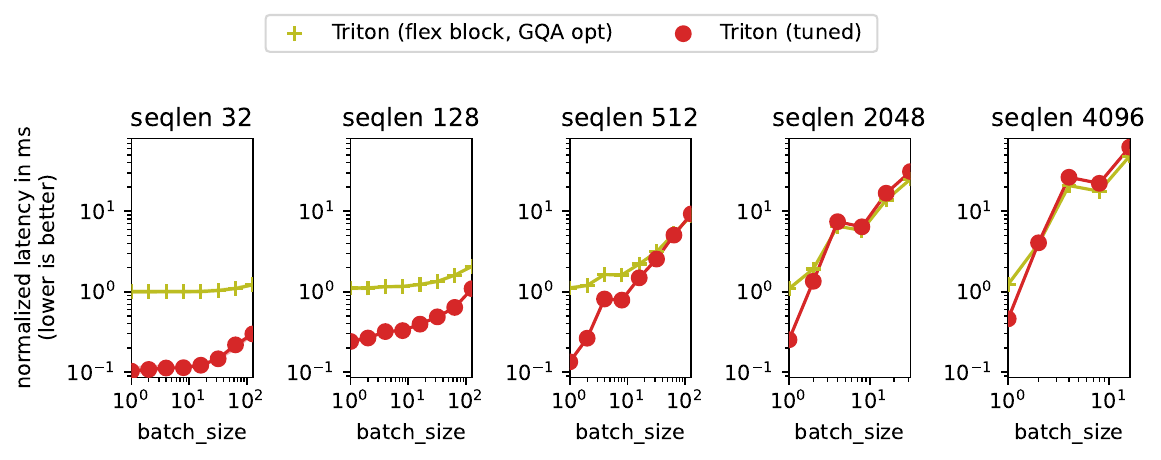}
		\vspace*{-2em}
		\caption{H100, sorted by max. sequence length}
		\label{subfig:eval-tuned-h100}
	\end{subfigure}%
	\begin{subfigure}[t]{.5\textwidth}
		\includegraphics[width=\linewidth]{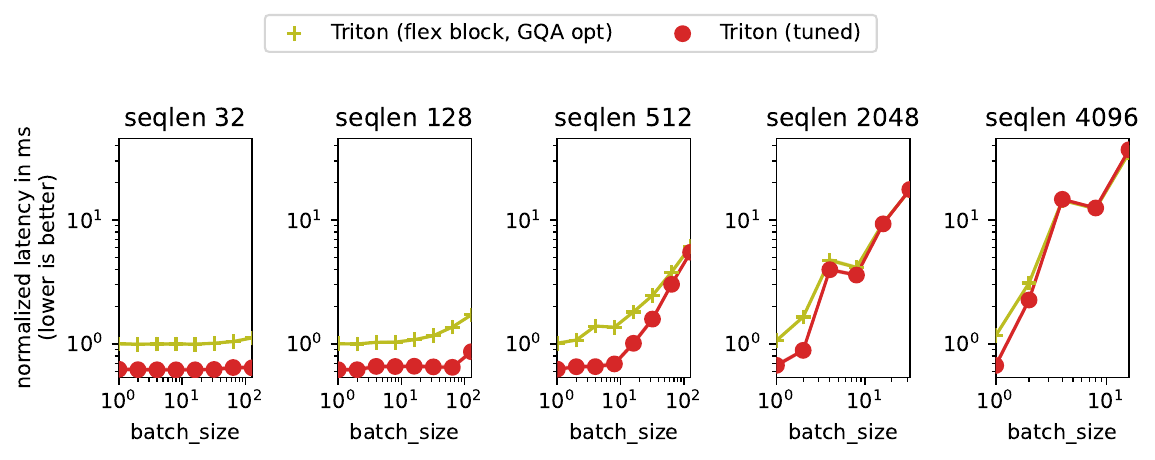}
		\vspace*{-2em}	
		\caption{MI300, sorted by max. sequence length}
		\label{subfig:eval-tuned-mi300}
	\end{subfigure}%
	\vspace*{-0.5em}
	\caption{Comparing performance of the optimized QGA kernel with and without tuning for prefill sizes.
	}
	\label{fig:eval-tuned}
	\vspace*{-1em}
\end{figure*}

\section{Performance Evaluation}

We evaluate the kernels using a two-track approach: Measuring incremental performance changes with a micro-benchmark suite, followed by comprehensive end-to-end testing. 
Our evaluation aims to answer the following research questions:
\begin{enumerate}
	\item \textit{How big is the performance impact of the individual optimization steps (c.f. \autoref{sec:kernel-implementation})?}
	\item \textit{Can autotuning, in the form of simple heuristics, improve the performance further (c.f. \autoref{sec:autotuning})?}
    \item \textit{What is the end-to-end performance of the whole inference server?}
    \item \textit{What are the effects of CUDA/HIP graphs vs. dynamic JIT compilation?}
\end{enumerate}

\subsection{Methodology}
We run our evaluation on two GPUs, the NVIDIA H100-80GB and the AMD MI250-128GB. We selected these two GPUs because they utilize the same technology nodes (\SI{5}{\nano\meter} manufactured by TSMC), represent the two major HW vendors, and because of their popularity and availability.

For the micro-benchmarks, we base our kernel parameters on the Llama3-8B LLM architecture~\cite{MetallamaLlama318BInstructHugging2024} (128 head size, 32 query heads, and 8 KV heads) and vary sequence lengths and batch sizes based on real-world samples. The sequences contained within a batch have variable lengths, as is often the case in real-world online inference scenarios. For every measurement in the micro-benchmarks, the kernel is warmed up with 20 iterations, and we take the mean of the 100 following iterations as the result. 


We run end-to-end experiments using the benchmark suite included with vLLM~\cite{vllm}, on both NVIDIA H100 and AMD MI300 GPUs. 
For these measurements, we disable prefix caching of vllm, since we want to evaluate kernel improvements and not prompt redundancies. 
We use random data and ignore the end-of-sequence token of the model. The number of warmup and measurement iterations is kept at their default values of 10 and 30, respectively.

\subsection{Evaluation of the Optimization Steps}
\label{subsec:microbench-eval}

To evaluate the individual optimization steps, we used micro-benchmarks to ensure precise measurements of this single component. 
The \autoref{subfig:eval-step1-h100} shows four different kernel implementations of paged attention evaluated on an H100. The \texttt{flash\_attn} refers to the state-of-the-art library Flash Attention 3~\cite{ShahFlashAttention3FastAccurate2024}, while the other three implementations are our Triton implementations. \texttt{Triton (naive)} is our baseline implementation from \autoref{par:baseline}, \texttt{Triton (GQA opt.)} is the version described in \autoref{par:q_blocks}, and \texttt{Triton (parallel tiled)} refers to the optimization explained in \autoref{par:par_tiled_softmax}. The very same implementations are shown \autoref{subfig:eval-step1-mi300}, for the AMD MI300 GPU. 
However, for AMD GPU there is no competitive paged attention implementation besides ours, hence there are only three Triton implementations depicted. 
In the subfigures \autoref{subfig:eval-step1-h100} and \autoref{subfig:eval-step1-mi300}, the individual plots represent batches with the maximum sequence length as denoted on the top, and the batch size is shown on the x-axis. The y-axis shows the latency. 
The leftmost value of the baselines is used to normalize all latency values of the micro-benchmarks.  

If we examine the orange curves, which represents the naive implementation, we see that they are nearly an order of magnitude slower than FlashAttention across all sequence lengths. The optimization for better data locality, especially with GQA models (shown as the purple curve), shows improvements over the naive implementation for small sequence lengths and small batch sizes. Sometimes, it is even faster than FlashAttention. 
But, it does not bring significant benefits for sequence lengths larger than 500-1000 tokens. The optimization of parallel tiled softmax appears to increase the performance even less. 

However, if we plot the very same data not by sequence lengths, but by the different compositions of the batch, as done in subfigures \ref{subfig:eval-step1-h100-typewise} and \ref{subfig:eval-step1-mi300-typewise}, we see a totally different behavior. For these figures, we aggregated the batch size times the sequence lengths in the batch on the x-axis and divided the plots by the percentage of decode-only requests in the batch: 0\%, 50\%, and 100\%. Since vLLM is always prioritizing decode requests, batches with a high share of decode requests are common. 
As shown in subfigures \ref{subfig:eval-step1-h100-typewise} and \ref{subfig:eval-step1-mi300-typewise}, the GQA optimized kernel version has its strength primarily in handling prefill-heavy batches, which is expected, since prefill is compute-bound. The memory-bound decode phase requires the greater parallelization of the parallel tiled softmax implementation. As depicted by the right most subplot of \autoref{subfig:eval-step1-h100-typewise}, the blue curve exhibits nearly the same performance as FlashAttention. For very long decodes, the parallel tiled softmax also outperforms the GQA optimized kernel variant, as shown in \autoref{subsec:end2end-eval}. 

In \autoref{fig:eval-step2} we compare the flexible block approach described in \autoref{par:adjustable-tiles}. 
The two best approaches of \autoref{fig:eval-step1} are also presented in this figure for comparison. Since it has proven more insightful, we show here only the plots sorted by decode share. As can be seen and is expected, the \texttt{flex block} versions both outperform their respective comparable implementations of the previous figure. 

\subsection{Evaluation of our Autotuning Approach}
\label{subsec:tune-eval}

Next, we evaluate our approach to autotuning. Our preliminary tests showed that autotuning mainly affects the prefill phase; therefore, \autoref{fig:eval-tuned} shows only evaluations for prefill-heavy batches, again plotted by sequence length. 
For the \texttt{tuned} version in \autoref{fig:eval-tuned}, we followed the flow described in \autoref{sec:autotuning} and decided to settle on a very simple decision tree, based on the individual autotuning results. The simple decision tree is shown in Listing~\ref{lst:eval-tune-decision-tree}. 
As can be seen, this limited autotuning approach further reduces the kernel latency for short (up to $9.8\times$ on H100) and medium prompts (up to 75\%) on both platforms. 
To not exceed the scope of this paper, we do not evaluate more complex heuristics since it would require more complex overhead and trade-off analyses. However, already this limited heuristics demonstrated to be beneficial and is also used in the end-to-end evaluation.

\begin{figure*}
\vspace*{-1em}
\centering
	\captionsetup[subfigure]{justification=centering}
\begin{subfigure}[t]{.49\textwidth}
    \includegraphics[width=\linewidth, trim={2 2 2 2},clip=true]{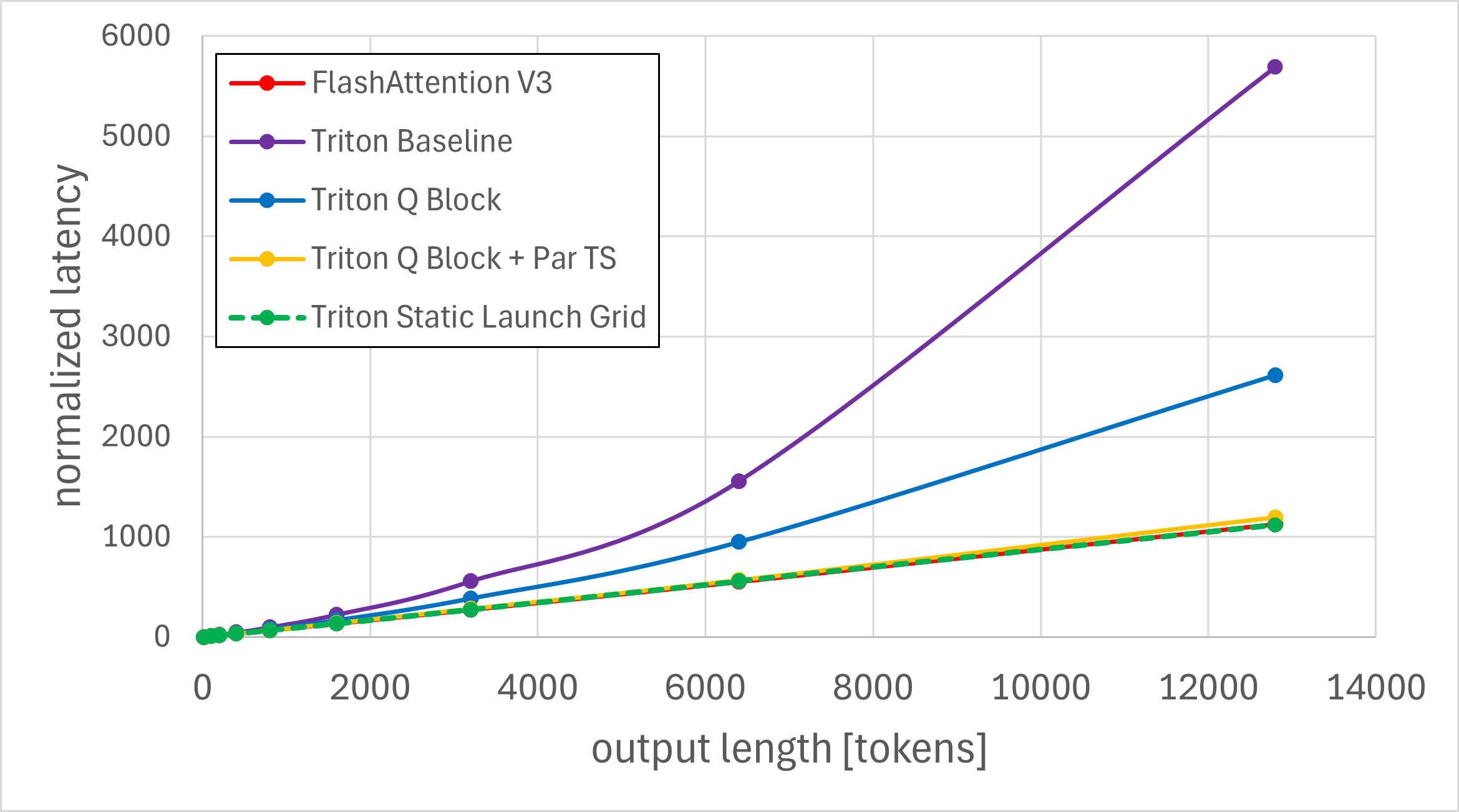}
    \vspace*{-2em}
    \caption{H100}
    \label{subfig:eval-latency-h100}
\end{subfigure}%
\hfill
\begin{subfigure}[t]{.49\textwidth}
    \includegraphics[width=\linewidth, trim={2 2 2 2},clip=true]{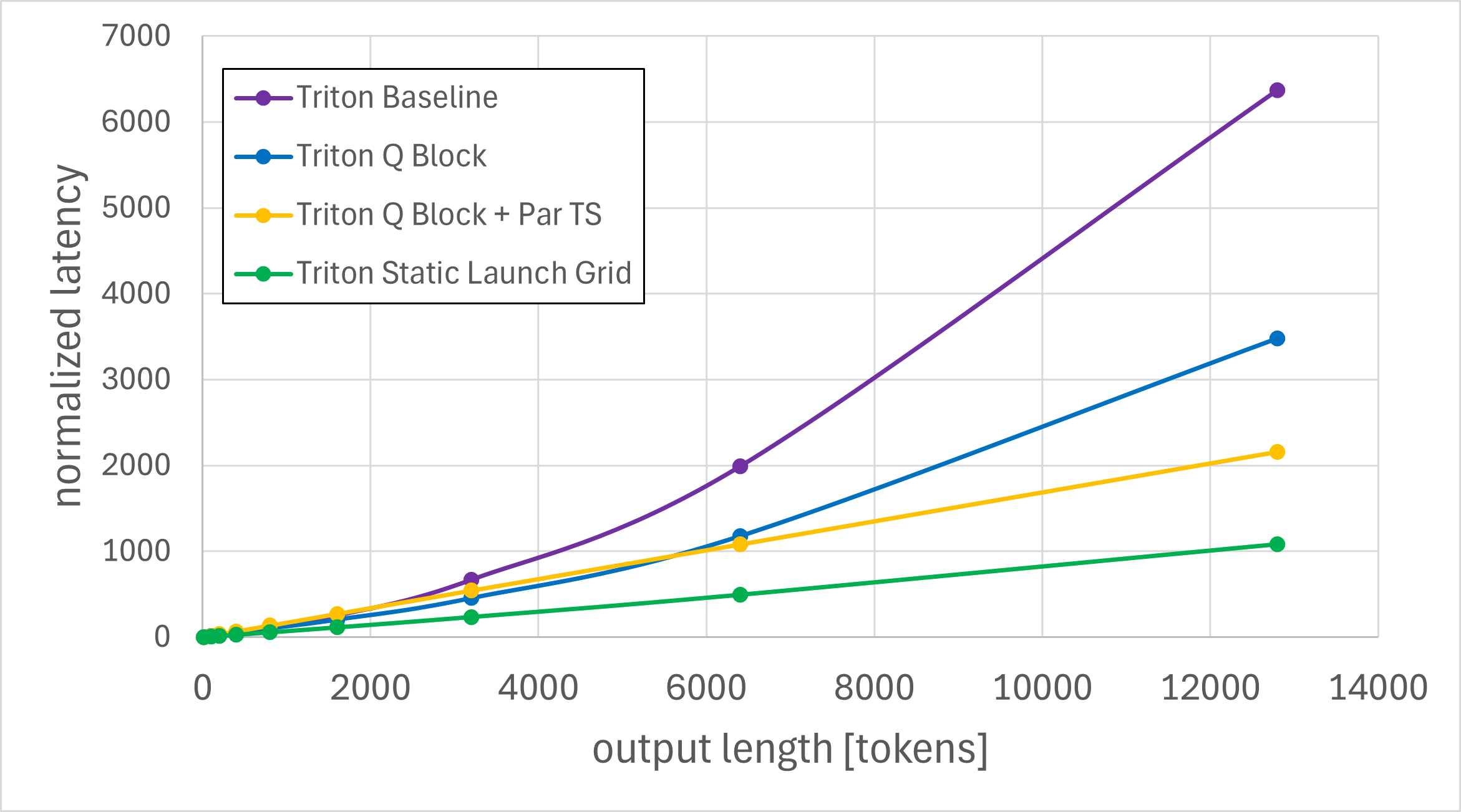}
    \vspace*{-2em}
    \caption{MI300}
    \label{subfig:eval-latency-mi300}
\end{subfigure}%
    \vspace*{-0.5em}
    \caption{
    vLLM latency benchmark results for Llama-3.1-8B with batch size 1 and input length of 500 tokens.
    }
    \label{fig:latency_results}
    \vspace*{-1em}
\end{figure*}

\subsection{End-to-End Evaluations}
\label{subsec:end2end-eval}

%


Last, we evaluate our kernel and system improvments for a popular model and large requests end-to-end.
Figure~\ref{fig:latency_results} presents the results of the latency benchmark for the meta-llama/Llama-3.1-8B-Instruct model~\cite{MetallamaLlama318BInstructHugging2024}, using a batch size of 1 and a prompt length of 500 tokens, evaluated across varying numbers of generated output tokens. This test configuration enables a detailed analysis of the impact of the various optimizations discussed in Section~\ref{sec:kernel-implementation} and excludes scheduling decisions from affecting the measurements. 

The Triton Static Launch Grid-based kernel integrates the Q-Block and Parallel Tiled Softmax optimizations found in the other kernels, and additionally incorporates a static launch grid (see \autoref{par:static_launch_grid}), along with being executed with full CUDA/HIP-graphs enabled (see also \autoref{subsec:cuda-graphs-vllm}). All other experiments were run with partial HIP/CUDA-graphs.
Furthermore, the Triton Static Launch Grid kernel includes heuristics for tile- and segment-sizes found during autotuning (c.f. \autoref{subsec:tune-eval}). 

The benchmark results obtained on the H100, shown in \autoref{subfig:eval-latency-h100}, demonstrate a consistent improvement across the successively optimized kernels. In particular, the Parallel Tiled Softmax optimization shows a clear advantage for longer sequences 
, reducing the observed end-to-end latency by more than a factor of two compared to the kernel that only implements the Q-Block optimization, for an output length of 12,800 tokens. 
The Static Launch Grid-based kernel, executed with full CUDA Graphs enabled, further reduces latency by approximately 6\% for the same output length. 
These results suggest that larger reductions in latency can be expected as sequence lengths increase. For reference, Figure~\ref{fig:latency_results} also includes results from the same experiments conducted using FlashAttention V3. From the comparison, it is evident that in this setup, the performance of the Static Launch Grid-based kernel is comparable to that of FlashAttention V3. As \autoref{fig:latency_results} shows, the baseline implementation achieved only 19.7\% of the performance of FlashAttention3. Our first optimized Triton kernel (c.f. \autoref{app:triton-qblock}) increases this by $2.1\times$ to 49\%, while the last optimization --- the static launch grid, c.f. \autoref{par:static_launch_grid} --- achieves even 98.6\% -- 105.9\% versus FlashAttention3. 

The results for the MI300, shown on the right-hand side of Figure~\ref{fig:latency_results}, reveal a higher impact of the launch overhead on performance compared to the H100. The Triton Static Launch Grid with heuristics and full-HIP-graphs is up to $1.99\times$ faster than the Q Block and Par Ts version. 
This is evident from the higher latency of the Parallel Tiled Softmax kernel relative to the Q-Block kernel for shorter output lengths, as the former includes an additional launch of a reduction kernel (\autoref{par:par_tiled_softmax}). For longer sequences, however, the launch overhead becomes negligible v.s. the actual processing time, allowing the Parallel Tiled Softmax optimization to take effect. The impact of the launch overhead is even more apparent from the substantial performance improvement achieved by enabling full HIP Graphs (a key solution for mitigating launch overhead) in the Triton Static Launch Grid-based kernel, which reduces latency by about a factor of two across all output lengths compared to the Parallel Tiled Softmax kernel.
Combined, the three optimizations shown in \autoref{subfig:eval-latency-mi300} exhibit a speedup of $5.9\times$.



\section{Insights and Future Work}
\label{sec:insights}

Throughout the presented work, we 
describe 
how to develop a Triton-only state-of-the-art attention kernel. However, we also discovered insights that extend beyond our specific use case. 
For example, these lesson learned are applied to improve the kernels for mamba/\gls{ssm} layers in vLLM~\cite{KernelChunkalignedMamba2}.
We discuss these insights in the following section and also mention future work. 

\paragraph{Triton kernels need to be specific} 
Throughout our research, we experimented with kernel versions that would merge the three different kernels described in \autoref{sec:kernel-implementation}. In particular, we had kernels that would detect if it is a decode-only in the beginning and then would branch to either the prefill kernel (cf. \autoref{par:q_blocks}) or a version of decoding-optimized kernel (similar to the baseline described in \autoref{par:baseline}). The goal of these experiments was to reduce the number of kernels launched, thereby minimizing the control and software overhead associated with these kernel launches. However, we observed that the performance of these kernels drops by at least $2x$, by far outweighing the saved launch overhead latency (in the order of \SI{150}{\micro\second}). Subsequent analysis of the Triton \glspl{ir} revealed that the software pipelining did not really produce useful pipelines in these fused kernels. Hence, we concluded that Triton kernels always need to be written around one specific problem with a strong data-dependency within this problem, and we would rather \eigName{pay} for multiple kernel launches otherwise. 

\vspace*{-0.5em}
\paragraph{Usage of \texttt{tl.dot}}
The multiplication $QK^\top$, which is part of the attention computation in Equation~\ref{eq:attention}, can be implemented in Triton using either of the following approaches:
\begin{itemize}
\item Element-wise multiplication between $K$ and a broadcasted version of $Q$, followed by a summation (reduction): \texttt{\small tl.sum(K * Q[:, None], axis=0)}.
\item A matrix multiplication instruction: \texttt{\small tl.dot(Q, K)}.
\end{itemize}
This approach also extends to multiplying the softmax output by $K$.

The second approach, using \texttt{\small tl.dot}, requires that the dimensions of the input matrices be multiples of specific MMA (Matrix Multiply-Accumulate) tile sizes (e.g., 8, 16, 32).
This often necessitates padding to satisfy these requirements. Despite this, it is generally preferred, as it almost always results in better performance.
This is because the compiler will map it directly to the MMA units, such as NVIDIA's Tensor Cores. In contrast, the element-wise multiplication followed by summation is often not recognized as matrix multiplication by the compiler.

\vspace*{-0.5em}
\paragraph{CUDA/HIP graphs do not always help}
We observe a common trend in the community to address the overhead of Tritons \gls{jit}- and launch-overhead (or also the overhead of other JIT frameworks, such as CUTLASS): If in doubt, use HIP/CUDA graphs. As discussed in \autoref{subsec:cuda-graphs-vllm}, CUDA/HIP graphs come with trade-offs. However, using such recorded graphs at the kernel/binary level forces the application to select \textit{exactly one} binary version of the Triton kernel, thereby sabotaging the philosophy behind a \gls{jit} language. Triton specializes kernels not only based on constants, but also on memory access patterns (e.g., if a stride is divisible by 16). Autotuning (cf. \autoref{sec:autotuning}) further adds to this diversity. Hence, we observed that in scenarios where a single kernel runtime is equal to the average launch overhead of Triton (in the order of \SI{200}{\micro\second}), HIP/CUDA graphs do not help in decreasing overall latency, unless a kernel is developed for this purpose, as described in \autoref{subsec:cuda-graphs-vllm}. 

\vspace*{-0.5em}
\paragraph{Future Work: Consider GPU-specific-parameters in kernel tuning}
Besides the platform-agnostic parameters like \texttt{num\_warps} (related to the number of threads) and \texttt{num\_} \texttt{stages} (related to the depths of the software pipeline),  newer Triton versions introduced GPU-specific parameters. For example, \texttt{num\_consumer\_groups} and \texttt{num\_buffers\_warp\_spec} for a concept called \eigName{Warp specialization} for Hopper and Blackwell GPUs~\cite{NVIDIACorporationNVIDIAH100Tensor2023, AutomaticWarpSpecialization}. Another example is \texttt{waves\_per\_eu} for AMD GPUs~\cite{OptimizingTritonKernels}. The parameters can be explored using the approach described in \autoref{sec:autotuning}, but remain outside the scope of this paper. 
However, in early experiments using these parameters in tuning showed an improvement of up to $80\%$ for some kernels. Related work in literature reports even improvements above $100\%$, for example, by using warp specialization in \texttt{flash\_attn3}~\cite{ShahFlashAttention3FastAccurate2024}. 

\paragraph{Future Work: Combining Autotuning with CUDA/HIP graphs}
CUDA/HIP graphs limit the flexibility of kernel-variant selection at runtime, as discussed above. Hence, it is also not possible to adapt a kernel block size and subsequently the launch grid of this kernel to the concrete number of tokens in a batch if using CUDA/HIP graphs. 
Consequently, to still be able to leverage the flexibility of autotuning, we have to move the decision trees, e.g., deciding the block size \textit{inside} a Triton kernel. Hence, the Triton binary remains the same, but it can adjust its loops accordingly, with some tweaking.
However, Triton loops do not support early exits using break or return statements. Hence, we have to insert instructions that do not have an impact (i.e., NOPs) to be able to realize the required masking (cf. \autoref{lst:attention_par_tiled_softmax}). In our experiments, this type of tuning yielded lesser performance improvements than transitioning to a static launch grid; therefore, this optimization remains future work. 


\section{Conclusions}

The democratization of AI depends on both the widespread use of LLM and AI models, but likewise on the flexibility and portability of AI deployments. Therefore, implementations of LLM solutions must allow for the selection of different hardware platforms. 
In this work, we argue that open-source solutions, such as Triton and vLLM, are key enablers for achieving this goal. 
We demonstrate that our Triton implementations of the paged attention kernel achieve state-of-the-art performance on NVIDIA and AMD GPUs, using the same source code. 
During our research, we learned lessons regarding the handling of the Triton memory hierarchy, improving data locality, auto-tuning to improve performance-portability, and balancing the integration of Just-in-Time compiled binaries in dataflow-only graphs, such as CUDA and HIP graphs. 
These investigated optimizations exhibit a total speedup of up to 589\% 
and we contributed our kernels and frameworks as open-source (\href{https://ibm.biz/vllm-ibm-triton-lib}{ibm.biz/vllm-ibm-triton-lib}). 
Finally, our resulting highly optimized kernels are the default attention backend in vLLM for AMD deployments.


\begin{acks}
We would like to thank our colleagues Chih-Chieh Yang and Sara Kokkila Schumacher for their always helpful discussions of Triton behavior and feedback to portable designs. 
\end{acks}

\pagebreak

\printbibliography


\clearpage

\appendix

\section{Triton Baseline Attention Kernel}
\label{app:triton-baseline}

\begin{listing}
    \captionof{listing}{Baseline Triton Attention Kernel.}
    \label{lst:attention_baseline}
    \begin{minted}
        [
        frame=lines,
        %framesep=2mm,
        %baselinestretch=1.2,
        %bgcolor=LightGray,
        fontsize=\scriptsize,
        linenos,
        ]
        {python}
@triton.jit
def kernel_attention(...):
    query_token_idx = tl.program_id(0)
    query_head_idx = tl.program_id(1)
    kv_head_idx = query_head_idx // num_queries_per_kv

    # determine sequence idx and prefix length for current token
    seq_idx = find_seq_idx(query_token_idx, ...)
    prefix_len = calc_prefix_len(seq_idx, query_token_idx, ...)

    # load Q for current query head
    #     for current token in current sequence
    Q = tl.load(...)

    # initialize tiled softmax tensors
    res, max, expsum = ...

    # iterate through tiles to attend current token to all
    #     previous tokens
    num_tiles = (prefix_len + BLOCK_SIZE - 1) // BLOCK_SIZE
    for j in range(0, num_tiles):
        # load j-th K and V tiles from KV cache
        #     for current KV head and current sequence
        K_j = tl.load(...)
        V_j = tl.load(...)

        # calculate tiled softmax
        #     update res, max, and expsum
        attn_result, max, expsum =
            tiled_attn(Q, K_j, V_j, attn_result, max, expsum)

    # store result
    attn_result = attn_result / expsum
    tl.store(..., attn_result)

# prefill
kernel_attention[(tot_query_length, num_query_heads,)](...)

# decode
kernel_attention[(num_seqs, num_query_heads,)](...)
    \end{minted}
\end{listing}

Listing~\ref{lst:attention_baseline} presents a high-level, simplified representation of our
baseline implementation of an attention kernel in Triton. It assumes that $Q$, $K$, and $V$
have already been computed prior to the kernel launch and stored in the KV cache. The KV cache in vLLM is accessed
through a block table (analogous to a page table), although this is not directly discussed here, except
for references to the parameter \texttt{\small BLOCK\_SIZE}, which defines the maximum number of tokens stored in
a single KV cache block.

As can be seen from lines $36$ to $40$, the kernel is launched separately for sequences being in prefill and decode phases.
In the launch grids, \texttt{\small num\_seqs} denotes the number of batched sequences, while
\texttt{\small tot\_query\_length} refers to the total number of tokens
across these sequences for which attention has to be calculated (i.e., the sum of the query lengths for all batched sequences).
For prefill attention, the latter value is obtained by summing
the prompt lengths for all sequences, whereas for decode attention,
it simply equals the number of sequences in the batch.

The launch grids show that a separate program instance is launched for each
combination of a query token and a query head. The program execution begins
by identifying the KV head corresponding to the given query head (line $5$) and
retrieving the sequence index (using a binary search on a tensor storing the accumulated query lengths for all sequences in the batch)
and prefix length for the given query token (lines $8$ and $9$). Next, the corresponding $Q$ matrix is loaded (line $13$).
Lines $15$ to $30$ implement the tiled softmax computation, following
the approach for tiled softmax outlined in section~\ref{par:core_concepts}. The tile size equals \texttt{\small BLOCK\_SIZE} (i.e., the KV cache block size).

\section{Triton Attention Kernel Optimized for Prefill and GQA}
\label{app:triton-qblock}

\begin{listing}
    \captionof{listing}{Attention Kernel Optimized for Prefill and QGA.}
    \label{lst:attention_q_block}
    \begin{minted}
        [
        frame=lines,
        %framesep=2mm,
        %baselinestretch=1.2,
        %bgcolor=LightGray,
        fontsize=\scriptsize,
        linenos,
        ]
        {python}
@triton.jit
def kernel_attention(...):
    q_block_idx = tl.program_id(0)
    kv_head_idx = tl.program_id(1)
    query_head_idx = kv_head_idx * num_queries_per_kv +
        tl.arange(0, num_queries_per_kv)

    # determine sequence idx and prefix length for current token
    seq_idx = find_seq_idx(q_block_idx, ...)
    max_prefix_len = calc_prefix_len(seq_idx, q_block_idx, ...)

    # load Q for current query head
    #     for current token in current sequence
    Q = tl.load(...)

    # initialize tiled softmax tensors
    attn_result, max, expsum = ...

    # iterate through tiles to attend current token to all
    #     previous tokens
    num_tiles = (max_prefix_len + BLOCK_SIZE - 1) // BLOCK_SIZE
    for j in range(0, num_tiles):
        # load j-th K and V tiles from KV cache
        #     for current KV head and current sequence
        K_j = tl.load(...)
        V_j = tl.load(...)

        # calculate tiled softmax
        #     update attn_result, max, and expsum
        attn_result, max, expsum =
            tiled_attn(Q, K_j, V_j, attn_result, max, expsum)

    # store result
    attn_result = attn_result / expsum
    tl.store(..., attn_result)

# prefill
kernel_attention[(tot_num_q_blocks, num_kv_heads,)](...)

# decode
kernel_attention[(num_seqs, num_kv_heads,)](...)
    \end{minted}
\end{listing}

Listing~\ref{lst:attention_q_block} presents the Triton implementation of the Q-Block based optimized attention kernel, using a similar high-level and simplified representation
as with the baseline kernel. A key difference lies in the launch grid which now includes the total number of Q Blocks across all sequences in the batch, combined with the number
of KV heads (line $38$). For decode, the total number of Q Blocks equals the number of batched sequences (line $41$). Furthermore, now the query head indices are derived from
the KV head (lines $5$-$6$). The sequence index is determined in a similar way using a binary search on a tensor storing the accumulated number of Q Blocks for all sequences in the batch (line $9$).

\section{Triton Attention Kernel Using Parallel Tiled Softmax}
\label{app:triton-par-ts}

\begin{listing}
    \captionof{listing}{Kernel Supporting Parallel Tiled Softmax.}
    \label{lst:attention_par_tiled_softmax}
    \begin{minted}
        [
        frame=lines,
        %framesep=2mm,
        %baselinestretch=1.2,
        %bgcolor=LightGray,
        fontsize=\scriptsize,
        linenos,
        ]
        {python}
@triton.jit
def kernel_attention_par_ts(...):
    q_block_idx = tl.program_id(0)
    kv_head_idx = tl.program_id(1)
    segm_idx = tl.program_id(2)

    query_head_idx = kv_head_idx * num_queries_per_kv +
        tl.arange(0, num_queries_per_kv)

    # determine sequence idx and prefix length for current token
    seq_idx = find_seq_idx(q_block_idx, ...)
    max_prefix_len = calc_prefix_len(seq_idx, q_block_idx, ...)

    # load Q for current query head
    #     for current token in current sequence
    Q = tl.load(...)

    # initialize tiled softmax tensors
    segm_res, segm_max, segm_expsum = ...

    # iterate through tiles within current segment
    num_tiles = (max_prefix_len + BLOCK_SIZE - 1) // BLOCK_SIZE
    for j in range(
            segm_idx * tiles_per_segment,
            min((segm_idx + 1) * tiles_per_segment, num_tiles),
    ):
        # load j-th K and V tiles from KV cache
        #     for current KV head and current sequence
        K_j = tl.load(...)
        V_j = tl.load(...)

        # calculate tiled softmax
        #     update segm_res, segm_max, and segm_expsum
        segm_result, segm_max, segm_expsum =
            tiled_attn(Q, K_j, V_j, segm_result, segm_max, segm_expsum)

    # store segment results
    tl.store(..., segm_result)
    tl.store(..., segm_max)
    tl.store(..., segm_expsum)


@triton.jit
def reduce_segments(...):
    query_token_idx = tl.program_id(0)
    query_head_idx = tl.program_id(1)

    segm_idx = tl.arange(0, num_segments)

    segm_result = tl.load(...)
    segm_max = tl.load(...)
    segm_expsum = tl.load(...)

    # calculate overall result by merging and rescaling segment results
    attn_result = merge_segments(segm_results, segm_max, segm_expsum)

    tl.store(..., attn_result)


# decode
kernel_attention_par_ts[(num_seqs, num_kv_heads, num_segments)](...)
reduce_segments[(num_seqs, num_query_heads)](...)


    \end{minted}
\end{listing}

To extract enough parallelism during decode attention, the iterative processing of tiled softmax within the body of the \texttt{\small for} loop at line $22$ in Listing~\ref{lst:attention_q_block}  can be parallelized. 
Listing~\ref{lst:attention_par_tiled_softmax} presents the corresponding Triton implementation. In this version, a three-dimensional launch grid is used to launch the attention kernel, with the number of segments per Q Block and KV head combination forming the third dimension (line $61$). The segment index (line $5$), assigned to each program instance, determines which subset of tiles will
be processed iteratively within that program instance (lines $21$ to $26$). Once the attention computation for a segment is completed, the intermediate results are then stored in
memory (lines $37$ to $40$). After the first kernel finishes, the reduction kernel is launched to compute the final attention output from the segment-level results.

\end{document}